\documentclass{article}

    \PassOptionsToPackage{numbers, compress}{natbib}
    \usepackage[final]{neurips_2022}

\usepackage[utf8]{inputenc} % allow utf-8 input
\usepackage[T1]{fontenc}    % use 8-bit T1 fonts
\usepackage{hyperref}       % hyperlinks
\usepackage{url}            % simple URL typesetting
\usepackage{booktabs}       % professional-quality tables
\usepackage{amsfonts}       % blackboard math symbols
\usepackage{nicefrac}       % compact symbols for 1/2, etc.
\usepackage{microtype}      % microtypography
\usepackage{xcolor}         % colors
\usepackage{colortbl}

\usepackage[pdftex]{graphicx}
\usepackage{setspace,latexsym,amsmath,amssymb,amsthm,bm}
\usepackage[font=small,labelfont=bf]{caption}
\usepackage[draft]{commands}
\usepackage{subcaption}
\usepackage{import}
\usepackage{dirtytalk}
\usepackage{algorithm,algorithmic}
\usepackage{subcaption}
\usepackage{changepage}
\usepackage{enumitem}
\usepackage{listings}

\colorlet{bestcolor}{gray!30}
\newcommand\shade{\cellcolor{bestcolor}}

\title{A Spectral Approach to Item Response Theory}

\author{%
  Duc Nguyen\\
  Department of Computer and Information Science\\
  University of Pennsylvania \\
  \texttt{mdnguyen@seas.upenn.edu} \\
  \And
  Anderson Y. Zhang \\
  Department of Statistics and Data Science \\
  University of Pennsylvania \\
  \texttt{ayz@wharton.upenn.edu} \\
}

\begin{document}

\maketitle
\vspace{-1em}
\begin{abstract} The Rasch model is one of the most fundamental models in \emph{item response theory} and has wide-ranging applications from education testing to recommendation systems. In a universe with $n$ users and $m$ items, the Rasch model assumes that the binary response $X_{li} \in \{0,1\}$ of a user $l$ with parameter $\theta^*_l$ to an item $i$ with parameter $\beta^*_i$ (e.g., a user likes a movie, a student correctly solves a problem) is distributed as $\Pr(X_{li}=1) = 1/(1 + \exp{-(\theta^*_l - \beta^*_i)})$. In this paper, we propose a \emph{new item estimation} algorithm for this celebrated model (i.e., to estimate $\beta^*$). The core of our algorithm is the computation of the stationary distribution of a Markov chain defined on an item-item graph. We complement our algorithmic contributions with finite-sample error guarantees, the first of their kind in the literature, showing that our algorithm is consistent and enjoys favorable optimality properties. We discuss practical modifications to accelerate and robustify the algorithm that practitioners can adopt. Experiments on synthetic and real-life datasets, ranging from small education testing datasets to large recommendation systems datasets show that our algorithm is scalable, accurate, and competitive with the most commonly used methods in the literature.
\end{abstract}
\vspace{-1em}

% TODO:
% \begin{itemize}
% \item Explanation and comparisons to the BTL model
% \end{itemize}

\section{Introduction}

Item response theory (IRT) is the study of the relationship between latent characteristics (a student's ability versus a test's difficulty or a user's taste versus a movie's features) and the manifestations of these characteristics (a student's performance on a test or a user's rating of a movie). Originally developed by the psychometric community \cite{rasch1960studies,van1997handbook}, item response theory has been applied to diverse settings such as education testing \cite{lord2012applications}, crowdsourcing \cite{whitehill2009whose}, recommendation systems \cite{chen2005personalized}, finance \cite{schellhorn2013using} and marketing research \cite{brzezinska2016latent}. 

One of the most fundamental models in IRT is the Rasch model \cite{rasch1960studies}. It models the \emph{binary response $X_{li} \in \{0,1\}$} of user $l$ with latent parameter $\theta^*_l \in \R$ to item $i$ with latent parameter $\beta^*_i\in \R$ by
\begin{equation}\label{eqn:rasch-prob}
 \Pr(X_{li} = 1) = \frac{1}{1+\exp{-(\theta^*_l - \beta^*_i)}} \,.
\end{equation}
For example, in education testing, $\theta^*_l$ corresponds to the ability of student $l$, $\beta^*_i$ the difficulty of problem $i$ and $X_{li}=1$ if the student correctly solves the problem. Binary response data has grown abundantly in modern domains: Netflix famously switched from a 5-star rating system to a binary like/dislike feedback system, data on students' engagement and performance grows significantly as education moves online during the pandemic. 

Traditionally, the goal of estimation under the Rasch model is to \emph{recover the item parameters $\beta^*$}. In education testing, an estimate of the item parameters can be used to calibrate scores across different versions of a test. In recommendation systems, the item parameters can be used to produce a ranking over the items. In general, estimation is challenging under the Rasch model because for each user and item pair, we only get a single observation or none in the case of missing data. 

Joint maximum likelihood estimate (JMLE) is one of the earliest methods developed for the estimation problem \cite{andersen1973conditional,fischer1981existence,haberman1977maximum,hambleton1991fundamentals}. It estimates both the user and item parameters by maximizing the joint likelihood function using an alternating maximization algorithm. While efficient, JMLE is known to be \emph{inconsistent} (that is, even as $n\rightarrow \infty$, JMLE does not recover $\beta^*$) when the number of items is finite \cite{andersen1973conditional,ghosh1995inconsistent} (e.g., Figure \ref{fig:sfig1}). Intuitively, this is because there are many nuisance user parameters to a finite number of item parameters. As a result, JMLE is mostly used for prelimary parameter estimation and researchers have developed other solutions to address the inconsistency problem, broadly consisting of 3 approaches as follows.

The first approach is marginal maximum likelihood estimate (MMLE) \cite{basu2011elimination}. The statistician first specifies a prior distribution over the user parameters. The objective of MMLE is to maximize the marignal likelihood function which \emph{integrates out} the user parameters. In pratice, MMLE runs quite fast, handles missing data well and is reasonably accurate. However, its performance \emph{depends on the accuracy of the prior distribution}. If misspecified, MMLE may produce inaccurate estimates (e.g., Figure \ref{fig:sfig2}). Model selection is thus a crucial procedure when applying MMLE to real data.

The second approach is conditional maximum likelihood estimate (CMLE) \cite{andersen1973conditional,fischer1981existence,hambleton1991fundamentals}. CMLE builds on the fact that under the Rasch model the total number of positive responses $s_l$ for each user $l$ is a sufficient statistic for the user parameter $\theta^*_l$. Instead of the joint likelihood function, CMLE maximizes the likelihood conditioned on $\{s_l\}_{l=1}^n$. Unlike JMLE, CMLE is \emph{statistically consistent without requiring any distribution assumptions about $\theta^*$}. For small datasets with no missing data, CMLE is quite accurate. However, it may incur \emph{high computational cost and numerical issues} on large datasets with many items and missing entries. Practioners have observed that CMLE often produces inaccurate estimates \cite{linacre, linacre2} in this regime (e.g., Figure \ref{fig:sfig3}).

The third approach, which our algorithm follows, uses \emph{pairwise information} that can be extracted from binary responses. Intuitively, if a user responds to two items, one negatively and one positively, we learn that the later is `better'. Following this intuition, previous authors \cite{garner2002eigenvector,choppin1982fully,saaty1987analytic} have designed spectral algorithms that first construct an item-item matrix and then compute its leading eigenvector. One common limitation of these methods is that the item-item matrix is \emph{assumed to be dense}. Therefore, these methods aren't directly extendable to large scale datasets in applications such as recommendation systems where the item-item observation is sparse. 

Furthermore, most theoretical guarantees for the above methods are asymptotic $(n\rightarrow \infty)$. However, having finite sample error guarantees is useful in real-life applications. For example, when we only observe a handful of responses to a new item, it is important to have an accurate estimate of the error over the item parameter. Asymptotic guarantees, on the other hand, are accurate mostly under a large sample size regime, and can be inaccurate in the data-poor regime.

\textbf{Our Contributions:} Motivated by known limitations of the existing methods, we propose a new, theoretically grounded algorithm that addresses these limitations and performs competitively with the most commonly used methods in the literature. More specifically:
\begin{itemize}
\item In Sections \ref{sect:algorithm} and \ref{sect:accelerated}, we describe the spectral algorithm and practical modifications -- an accelerated version of the original algorithm and a regularization strategy -- that allow the algorithm to scale to large real-life datasets with sparse observation patterns and alleviate numerical issues.
\item In Section \ref{sect:analysis}, we present \emph{non-asymptotic} error guarantees for the spectral method -- the first of their kind in the literature -- in Theorems \ref{thm:parameters-error-bound} and \ref{coro:parameters-error-bound-growing-m}. Notably, under the regime where $m$ grows, the spectral algorithm has optimal (up to a constant factor) estimation error achievable by any unbiased estimator (Theorem \ref{thm:cramer-rao-lower-bound-known-theta}). Under the challenging regime where $m$ is a constant or grows very slowly we show that the spectral algorithm is, unlike JMLE, \emph{consistent} (Corollary \ref{coro:consistency}).
\item In Section \ref{sect:experiments}, we present experiment results on a wide range of datasets, both synthetic and real, to show that our spectral algorithm is \emph{competitive} with the most commonly used methods in the literature, \emph{works off-the-shelf with minimal tuning and is scalable} on large datasets.
\end{itemize}

\subsection{Notations and Problem Formulation}
As briefly described before, in a universe of $n$ users and $m$ items, each user $l$ has a latent parameter $\theta^*_l \in \R$ and each item $i$ has latent parameter $\beta^*_i \in \R$. The reader may recognize that there is a fundamental identifiability issue associated with the Rasch model pertaining to translation. That is, $\{\theta^*, \beta^*\}$ and $\{\theta^* + \alpha \mb 1_n, \beta^* + \alpha\mb 1_m \}$ describe the same model for any $\alpha \in \R$. For this reason, we impose a normalization constraint on the item parameters ${\beta^*}^\top \mb 1_m = 0$. We consider the fixed range setting where $\beta^*_i \in [\beta^*_{\min}, \beta^*_{\max}] \,\forall i\in[m]$ for some constants $\beta^*_{\min}$, $\beta^*_{\max}$. Similarly, we assume that $\theta^*_l \in [\theta^*_{\min}, \theta^*_{\max}]$ for some constants $\theta^*_{\min}, \theta^*_{\max}$ \footnote{The bounded range assumption is a common one in the literature on the Rasch model. Intuitively, it eliminates the presence of items that are always repsonded positively to (or negatively to) and users who only responds positively (or negatively) that leads to parameter unidentifiability \cite{haberman2004joint}.}. The observed data is $X \in \{0, 1, *\}^{n\times m}$ where $*$ denotes missing data and for entries where $X_{li}\neq *$, $X_{li}$ is independently distributed per Equation (\ref{eqn:rasch-prob}). Let $A \in \{0,1\}^{n\times m} $ denote the assignment matrix where $A_{li} = 1$ if user $l$ responds (either negatively or positively) to item $i$ and $0$ if user $l$ does not respond to item $i$ (i.e., $X_{li} = *$). Define $B = A^\top A$, i.e., $B_{ij}$ is the number of users who respond to both items $i,j$. The goal of item estimation is to obtain an estimate $\beta$ from the observed data $X$ and the metric of interest is the $\ell_2$ error, $\lVert \beta - \beta^* \rVert_2$.

\section{The Spectral Estimator}\label{sect:algorithm}

In this section we describe our spectral algorithm which is summarized in Algorithm \ref{alg:spectral}. At a high level, the algorithm constructs a Markov chain defined on a graph whose vertices are the items and its transition probabilities are estimated using the observed user-item response data. The algorithm then computes the stationary distribution of this Markov chain and the estimate $\beta$ is obtained following a simple transformation. 

We first define, for each item pair $i, j$ and a fixed assignment $A$, a quantity which we term \emph{pairwise differential measurement}:
\begin{equation}\label{eqn:Yij}
Y_{ij} =  \sum_{l=1}^n A_{li} A_{lj} X_{li}(1-X_{lj}) \quad \forall i\neq j\in [m]\,.
\end{equation}
Intuitively, $Y_{ij}$ is the number of users who respond $1$ to $i$ and $0$ to $j$.
Given the pairwise differential measurements, consider a Markov chain $P \in [0,1]^{m\times m}$ whose transition probabilities are defined as follows:
\begin{equation}\label{eqn:emp-markov-chain-def}
P_{ij} = \begin{cases} \frac{1}{d} Y_{ij} &\text{ if $i \neq j$}\\
1 - \sum_{k\neq i} \frac{1}{d} Y_{ik} &\text{ if $i = j$}
\end{cases}\quad ,
\end{equation}
where $d$ is a sufficiently large normalization factor chosen such that the resulting pairwise transition probability matrix does not contain any negative entries. Typically, $d = O(\max_{i\in [m]} \sum_{k\neq i} B_{ik})$. The algorithm then computes the stationary distribution $\pi$ of the Markov chain (e.g., using power iteration) and recover $\beta$ using a truncated log transformation step. The truncated transformation is used to facilitate the resulting theoretical analysis. The statistician could use any reasonable estimate of $\beta^*_{\max} - \beta^*_{\min}$ and incur little impact on practical performance of the algorithm. In real-life datasets, the constructed Markov chain is often sparse (not every pair of items has non-zero pairwise differential measurements). Practicioners could take advantage of this sparsity to speed up the computation of the stationary distribution such as by using sparse matrix-vector multiplication subroutines.

\begin{algorithm}[]
\caption{Spectral Estimator}
\hspace*{\algorithmicindent} \textbf{Input: } User-item binary response data $X\in \{0, 1, *\}^{n\times m}$. \\
\hspace*{\algorithmicindent} \textbf{Output: } An estimate of the item parameters $\beta = [\beta_1, \ldots, \beta_m]$.\\
\vspace{-0.75em}
\begin{algorithmic}[1]
    \STATE Construct a Markov chain $P$ per Equation (\ref{eqn:emp-markov-chain-def}).
    \STATE Compute the stationary distribution of $P$:\\
           \quad Initialize $\pi^{(0)} = [\frac{1}{m},\ldots, \frac{1}{m}]$.\\
           \quad For $t = 1, 2, \ldots$ until convergence, compute\\
           \quad $${\pi^{(t)}}^\top = \frac{{\pi^{(t-1)}}^\top P }{\lVert {\pi^{(t-1)}}^\top P \rVert_1 }.$$
    \STATE Compute $\bar\beta_i = \log\left(\max\left\{\pi_i, \frac{1}{me^{\beta^*_{\max} - \beta^*_{\min}}} \right\}\right)$ for $i \in [m]$.\\
    \STATE Return the normalized item parameters, i.e.,  $\beta = \bar\beta - \bar\beta^\top\mb 1/m$.\\
\end{algorithmic}
\label{alg:spectral}
\end{algorithm}

To understand the intuition behind our spectral algorithm, let us consider the following idealized Markov chain where the state transition probabilities are exact:
\begin{equation}\label{eqn:ideal-mc}
 P^*_{ij} = \begin{cases} \frac{1}{d} Y_{ij}^* &\text{for } i\neq j \\ 1 - \frac{1}{d} \sum_{k\neq i} Y^*_{ik} &\text{for } i = j \end{cases} \quad ,
\end{equation}
where $Y_{ij}^* =  \sum_{l=1}^n A_{li} A_{lj} \E[X_{li}(1-X_{lj})]$. For every pair $i, j$, given a sufficiently large number of users who respond to both items, $Y_{ij}$ will concentrate around $Y_{ij}^*$. Then, under an appropriately large scaling factor $d$, $P_{ij} \approx P^*_{ij}$ and the two Markov chains are `close'. This means that the stationary distribution of $P$ is also close to that of $P^*$. At the same time, the true item parameter $\beta^*$ is directly related to the stationary distribution of $P^*$. This relation is summarized by Proposition \ref{lem:reversibility}.
\begin{proposition}\label{lem:reversibility} Consider the idealized Markov chain described in Equation (\ref{eqn:ideal-mc}). The stationary distribution $\pi^*$ of $P^*$ satisfies $\pi^*_i = e^{\beta_i^*}/(\sum_{k=1}^m e^{\beta_k^*})$ for $i\in[m]$. 
\end{proposition}

Essentially Proposition \ref{lem:reversibility} states that $\pi^*$ is proportional to $e^{\beta^*}$. Thus $\beta^*$ can be recovered from $\pi^*$ up to a global normalization. Now, given a sufficiently large number of users, the empirical stationary distribution $\pi$ will be close to $\pi^*$ and naturally the obtained estimate $\beta$ is also close to $\beta^*$.

Readers who are familiar with the ranking from pairwise comparison literature might recognize the similarity between the spectral algorithm and Rank Centrality \cite{negahban2017rank} for parameter estimation under the Bradley-Terry-Luce model \cite{luce2012individual}. 
% Rank Centrality is an algorithm within a class of algorithms generally referred to as spectral ranking. 
Similarly to Rank Centrality, our algorithm constructs a Markov chain on the item-item graph and recovers parameter estimate from its stationary distirbution. In both cases, the Markov chain interpretation is motivated by the unique characteristics of the BTL and Rasch likelihood function. However, the Markov chain construction differs between our algorithm and Rank Centrality and so does the resulting analysis. 
% Generally speaking, the core algorithmic idea behind spectral type algorithms is the computation of the leading eigenvector of a matrix. In this sense, our algorithm instantiates the general spectral approach to the Rasch model estimation problem. 

\section{Theoretical Analysis}\label{sect:analysis}

In this section, we present the main theoretical contributions of the paper. Specifically, we obtain in Section \ref{sect:error-guarantees} two finite sample error bounds for two different regimes of $m$: where $m$ is a constant or grows very slowly and where $m$ grows at least logarithmically relative to $n$. In addition to our upper bounds, we show in Section \ref{sect:lower-bounds} a Cramer-Rao lower bound for the mean squared error of any unbiased estimator, establishing the optimality of the spectral algorithm under the the second regime. For the special case $m=2$, we show that the error rate obtained by the spectral algorithm is optimal up to a $\log$ factor.

\subsection{Finite Sample Error Guarantees}\label{sect:error-guarantees}
\textbf{Sampling Model:} Let us consider a random sampling model where for each user $l \in [n]$, each item $i \in [m]$ is independently shown to that user with probability $p$ (i.e., $\Pr(A_{li}=1) = p$). Once shown an item $l$, the user $i$ responds with $X_{li}$ distributed according to Equation (\ref{eqn:rasch-prob}).

Under this sampling model and the regime where \emph{$m$ is a constant or grows very slowly}, we obtain the following upper bound on the estimation error of the spectral algorithm which is, to the best of our knowledge, the first finite sample error guarantee for any consistent estimator under the Rasch model in the literature.
\begin{theorem}\label{thm:parameters-error-bound} Consider the sampling model described above. Suppose that $np^2 \geq C'\log m $ for a sufficiently large constant $C'$ then the output of the spectral algorithm statisfies
$$ \lVert \beta - \beta^* \rVert_2 \leq  \frac{C\sqrt{\max\{m, \log np^2 \} } }{\sqrt{np^2}}  $$
with probability at least $1- \min\{e^{-12m}, \frac{1}{(np^2)^{12}}\} - \exp{-C_1 np^2}$, where $C, C_1$ are constants.
\end{theorem}
As alluded to before in our algorithm description, the proof of Theorem \ref{thm:parameters-error-bound} uses Markov chain analysis and a central object is the idealized Markov chain $P^*$ with its stationary distribution $\pi^*$. The proof is rather long and involved so we defer the details to the supplementary materials and describe here the main idea. The starting point is a Markov chain eigen-perturbation bound (see Lemma \ref{lem:perturb-bound-mc}):
$$ \lVert \pi - \pi^* \rVert_2 \leq \frac{\lVert {\pi^*}^\top(P^* - P)\rVert_{2}  }{\mu^*(P^*) - \lVert P - P^* \rVert_{2}}\,,$$
where $\mu^*(P^*)$ is the \emph{spectral gap} of the idealized Markov chain. We then bound the numerator and the denominator separately. We will show under the setting of Theorem \ref{thm:parameters-error-bound} that
$$ \mu^*(P^*) - \lVert P - P^* \rVert_2 = \Omega\bigg(\frac{1}{d}\bigg)\quad \text{and}\quad  \lVert {\pi^*}^\top(P^* - P)\rVert_{2}  = O\bigg(\frac{\sqrt{\max \{m, \log np^2 \} }}{dm\sqrt{np^2}}\bigg) \,.  $$
% $$ \mu^*(P^*) - \lVert P - P^* \rVert_2 = \Omega\bigg(\frac{1}{d}\bigg) $$
% $$ \lVert {\pi^*}^\top(P^* - P)\rVert_{2}  = O\bigg(\frac{\sqrt{\max \{m, \log np^2 \} }}{dm\sqrt{np^2}}\bigg) \,. $$
Combining these bounds with the following relation gives us the desired error bound:
$$ \lVert \beta - \beta^* \rVert_2 = O\bigg(m \cdot \lVert \pi- \pi^*\rVert_2\bigg)\,. $$
As an immediate consequence of Theorem \ref{thm:parameters-error-bound}, we can also prove the consistency of the spectral algorithm under the constant $m$ regime. As mentioned previously, JMLE, one of the most well known methods in the Rasch modeling literature, is inconsistent in this regime.
\begin{coro}\label{coro:consistency} Consider the setting of Theorem \ref{thm:parameters-error-bound}. For a fixed $m$ and $p = 1$, the spectral algorithm is a consistent estimator of $\beta^*$. That is, its output $\beta$ satisfies
$\lim_{n\rightarrow \infty} \Pr(\lVert \beta - \beta^* \rVert_2 < \epsilon) = 1 \,, \forall \epsilon > 0\,.$
\end{coro}

\emph{Under the regime where $m$ grows}, we could sharpen the results of Theorem \ref{thm:parameters-error-bound}. Specifically, when the number of items shown to each user is sufficiently large, we improve by a $\sqrt{p}$ factor which can be significant when $p$ is small. This is summarized by the following theorem.
\begin{theorem}\label{coro:parameters-error-bound-growing-m} Consider the setting of Theorem \ref{thm:parameters-error-bound}. Assume further that $mp \geq C''\log n$ for a sufficiently large constant $C''$ then the output of the spectral algorithm statisfies
$$ \lVert \beta - \beta^* \rVert_2 \leq  \frac{C^*\sqrt m}{\sqrt{np}} $$
with probability at least $1- \exp{-C_2np^2} -2n^{-9}$, where $C^*, C_2$ are constants.
\end{theorem}

The reader may wonder why there would be a difference between the two regimes. Intuitively, when $m$ is a small constant, the distribution of the number of items shown to the users are not tightly concentrated. Some users are shown all of the items while some are shown only one. By design, our spectral algorithm uses pairwise differential measurements. This means that when a user responds to only one item, that information is not fully used. On the other hand, when $mp = O(\log n)$, the number of items shown to the users is concentrated (all users are shown approximately the same number of items) and more pairwise differential measurements are available. There is less information being under-utilized by the algorithm and it enjoys a tighter (in fact optimal) error rate.
\subsection{Cramer-Rao Lower Bound}\label{sect:lower-bounds}
In this section, we present complementary results to our finite error guarantees obtained in the previous section. Notably, under the regime where $m$ is allowed to grow with $n$, we show that the minimum mean squared error achievable by any unbiased estimator is \emph{no more than a constant factor} smaller than the upper bound for the spectral algorithm established in Theorem \ref{coro:parameters-error-bound-growing-m}. This optimality result is summarized by the following theorem.
\begin{theorem}\label{thm:cramer-rao-lower-bound-known-theta} Consider the sampling model described in in Section \ref{sect:error-guarantees}. Let $T$ be any unbiased estimator for the item parameters. Then the mean squared error of such estimator is lower bounded as
$$ \E\lVert T(X) - \beta^*\lVert^2_2 \,\geq \frac{cm}{np} \,, $$
where $T(X)$ is the output of the estimator $T$ when given data $X$ and $c$ is a constant.
\end{theorem}
Now note that under the settings of Theorem \ref{coro:parameters-error-bound-growing-m}, the output of the spectral algorithm also statisfies $\lVert \beta - \beta^* \rVert_2^2 = O(\frac{m}{np})$. To the best of our knowledge, this is the first non-asymptotic optimality result for any item estimation method under the Rasch model.

As noted before, when the number of items is constant, our error bound in Theorem \ref{thm:parameters-error-bound} incurs an additional $1/\sqrt p$ factor. We now argue that the upper bound obtained there may already be optimal in this challenging regime. Consider the special case when $m = 2$. Essentially, the goal is to estimate the \emph{difference of the two item parameters}. For a particular user $l$, suppose that we have no information about her parameter $\theta_l$ other than the bounded condition. If the user's responses consist of a single response (the response to one item is not observed) or that her responses to both items are identical (either both $0$ or $1$), then we learn little about the difference between the two items. We refer to these responses as `bad' responses. The relative difference between the items is only revealed if the user responds differently to the items. As noted in the description of our spectral algorithm, we refer to such information as \emph{pairwise differential measurements}. 

For the special case $m=2$, both CMLE and JMLE actually ignore `bad' responses. This is because in both algorithms, it has been shown that including bad responses in the respective objective likelihood function leads to parameter unidentifiability \cite{haberman2004joint}. As mentioned earlier, MMLE, requires an accurate prior distribution in order to obtain good estimate accuracy. This is not possible when we have no information about the user parameters. With the exception of MMLE, \emph{all estimation methods} that we are aware of in the literature only use pairwise differential measurements.

With these points considered, if we restrict our attention to the class of algorithms that use pairwise differential measurements, then the spectral algorithm indeed achieves the best possible (up to a log factor) estimation error.
\begin{theorem}\label{thm:cramer-rao-m2} Fix $m=2$ and consider the sampling model described in Section \ref{sect:error-guarantees}. Let $T$ be any unbiased estimator for the item parameters that only uses \emph{pairwise differential measurements}. Then the mean squared error of such estimator is lower bounded as
$$ \E\lVert T(X) - \beta^*\lVert^2_2 \,\geq \frac{c'}{np^2} \,, $$
where $T(X)$ is the output of the estimator $T$ when given data $X$ and $c'$ is a constant.
\end{theorem}
As seen from Theorem \ref{thm:parameters-error-bound}, the estimate produced by the spectral algorithm satisfies $\lVert \beta - \beta^* \rVert^2_2 = \tilde O(\frac{1}{np^2})$, establishing its near optimality.

\section{Practical Implementation Aspects}\label{sect:accelerated}
In this section, we discuss two important practical aspects that practicioners may consider when applying the spectral algorithm real-life datasets of which observation pattern may not correspond exactly to the sampling model described in Section \ref{sect:error-guarantees}. Firstly, we identify slow convergence as a problem encountered by the spectral algorithm when the data is skewed in the sense that some items are highly responded to by users while some items are rarely responded to. To address this issue, we propose an accelerated spectral algorithm that enjoys the same error guarantees as the original spectral algorithm but runs significantly faster in practice and suffers from fewer numerical issues. Secondly, we discuss regularization strategy when the spectral algorithm is applied to datasets with sparse observation patterns.

\textbf{Accelerating the Spectral Algorithm:} Recall that in the original spectral Algorithm \ref{alg:spectral}, we use a common normalization constant $d$ that generally scales as $O(\max_i \sum_{k\neq i} B_{ik})$. In practice, the distribution of the user-item assignment could be skewed such that some items are rarely responded to while some elicit many user responses. In such cases, the items with few responses will have few pairwise differential measurements $\ll d$. The induced Markov chain will contain large self-loops for these items. We observe that these large self-loops lead to a slower convergence when computing the stationary distribution and more numerical issues. This observation was also noted in \cite{agarwal2018accelerated} and we propose a similar solution to eliminate large self-loops that is to use a \emph{different normalizing factor for each vertex}. Consider the following modified Markov chain:
\begin{equation}\label{eqn:accelerated-mc}
\bar P_{ij} = \begin{cases} \frac{1}{d_i} Y_{ij} &\text{if } i \neq j\\
1 - \frac{1}{d_i} \sum_{k\neq i} Y_{ik} &\text{if } i = j
\end{cases}\quad ,
\end{equation}
where $Y_{ij}$ is defined in Equation (\ref{eqn:Yij}) and $\{d_i\}_{i=1}^m$ are appropriately chosen normalization factors such that the resulting transition probability matrix does not contain any negative entries. In our experiments, we choose $d_i = O(\sum_{k\neq i} B_{ik})$. The accelerated spectral algorithm computes the stationary distribution of the above modified Markov chain, and recovers the item parameters via a post-processing step. The algorithm is summarized in Algorithm \ref{alg:accelerated-spectral}.

\begin{algorithm}[]
\hspace*{\algorithmicindent} \textbf{Input: } User-item binary response data $X\in \{0,1,*\}^{n\times m}$. \\
 \hspace*{\algorithmicindent} \textbf{Output: } An estimate of the item parameters $\beta = [\beta_1, \ldots, \beta_m]$.\\
\vspace{-0.75em}
\begin{algorithmic}[1]
    \STATE Construct a \emph{modified} Markov chain $\bar P$ per Equation (\ref{eqn:accelerated-mc}).\\
    \STATE Compute the stationary distribution $\bar\pi$ of $\bar P$.\\
    \STATE Compute $\pi = (D^{-1}\bar \pi)/\lVert D^{-1}\bar\pi\rVert_1$ and $\bar\beta_i = \log\left(\max\left\{\pi_i, \frac{1}{me^{\beta^*_{\max} - \beta^*_{\min}}}\right\}   \right)$ where $D = \diag(d_1,\ldots, d_m)$.\\
    \STATE Return the normalized item parameters, i.e.,  $\beta = \bar\beta - \bar\beta^\top\mb 1/m$.\\
\end{algorithmic}
\caption{\emph{Accelerated} Spectral Estimator \label{alg:accelerated-spectral}}
\end{algorithm}

Interestingly, the accelerated algorithm produces essentially the same estimate as the original algorithm. Under some regularity conditions, there is a direct one-to-one relation between the stationary distribution $\pi$ obtained using the original Markov chain in Algorithm \ref{alg:spectral} and the stationary distribution $\bar\pi$ of the Markov chain parametrized by $\bar P$. This result is summarized in Theorem \ref{thm:equivalent-markov-chains}.
\begin{theorem}\label{thm:equivalent-markov-chains}
Consider the modified Markov chain $\bar P$ constructed per Equation (\ref{eqn:accelerated-mc}) and the original Markov chain $P$ constructed per Equation (\ref{eqn:emp-markov-chain-def}). Suppose that both $\bar P$ and $P$ admit unique stationary distributions $\bar\pi$ and $\pi$, respectively. Then
$$ \bar\pi_i = \frac{\pi_i d_i}{\sum_{k=1}^m\pi_k d_k  } \quad \forall i \in [m]\,,$$
where $d_i$ are the normalization factors in the construction of the modified Markov chain $\bar P$.
\end{theorem}
Assuming \emph{perfect numerical precision}, the two versions of the spectral algorithm output the same estimates. Therefore the guarantees in Theorems \ref{thm:parameters-error-bound} and \ref{coro:parameters-error-bound-growing-m} also apply to the accelerated spectral algorithm. However, in our experiments, we observe that the accelerated spectral algorithm converges much faster and suffers from fewer numerical issues than the original version on real-life datasets, leading to better performance overall. We thus use the accelerated version in our experiments.

\textbf{Regularization for Sparse Datasets:} In some real-life datasets, we observe that certain pairs of items have few pairwise differential measurements. Furthermore, the pairwise differential data is one-sided (e.g., users who respond to the two items always respond positively to one but negatively to the other). This could happen to pairs that have been shown to only few users. The existence of many such pairs may also introduce numerical issues and parameters unidentifiability. For example, when there is an item $i$ such that $Y_{ji} = 0 \, \forall j \neq i$, the stationary probability correpsonding to this item will be 0 and the item parameter estimate will be $-\infty$. As a solution, we propose adding regularization in the construction of the Markov chain. Specifically, for every pair of items $i, j$ such that $B_{ij} > 0$ redefine
$$ Y_{ij} = \sum_{l=1}^n A_{li}A_{lj} X_{li}(1-X_{lj}) + \nu \quad , $$
where $\nu$ is a small constant. In all of our experiments on real-life datasets, we use $\nu = 1$ and find that the regularization parameter requires little tuning. Regularization also ensures the uniqueness of the stationary distribution. So long as the graph underlying the Markov chain is connected, adding regularization ensures that no pairwise transition probability is 0. This makes the constructed Markov chain \emph{ergodic} and there is a unique stationary distribution \cite{norris1998markov}.

\section{Experiments}\label{sect:experiments}
In this section, we present empirical findings which support the practical value of our spectral algorithm. Our baselines are the most commonly used estimation algorithms in the literature: conditional maximum likelihood estimate (CMLE), marginal maximum likelihood estimate (MMLE) and joint maximum marginal likelihood (JMLE). Their open source implementation can be found online \cite{Sanchez_GIRTH_G_Item_2021}. We include the python implementation of our spectral algorithm in the supplementary materials.

\begin{figure}[]
\centering
\begin{subfigure}{.45\textwidth}
  \centering
  \includegraphics[height=40mm, width=60mm]{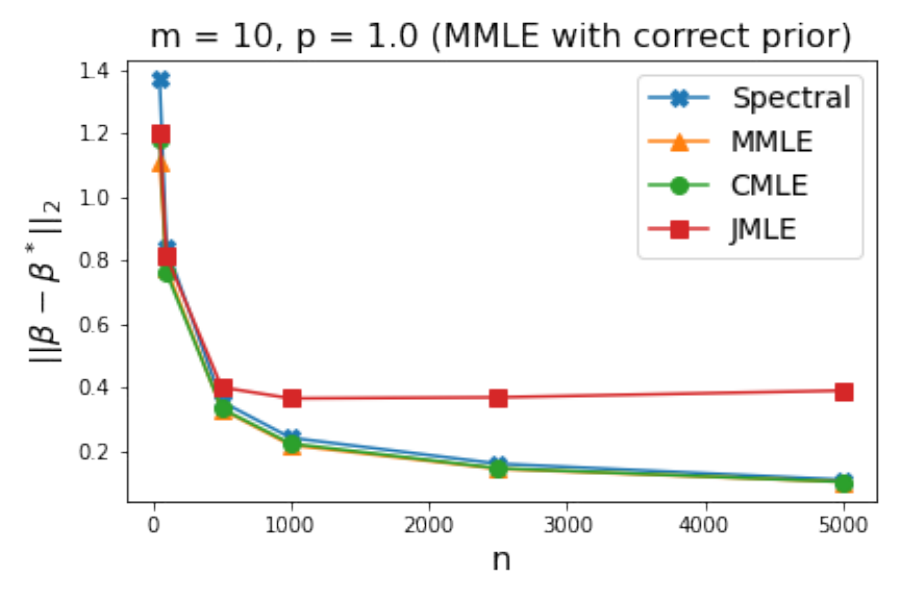}
  \caption{When the number of items is small, JMLE is known to produce inconsistent estimate.}
  \label{fig:sfig1}
\end{subfigure}
\begin{subfigure}{.45\textwidth}
  \centering
  \includegraphics[height=40mm, width=60mm]{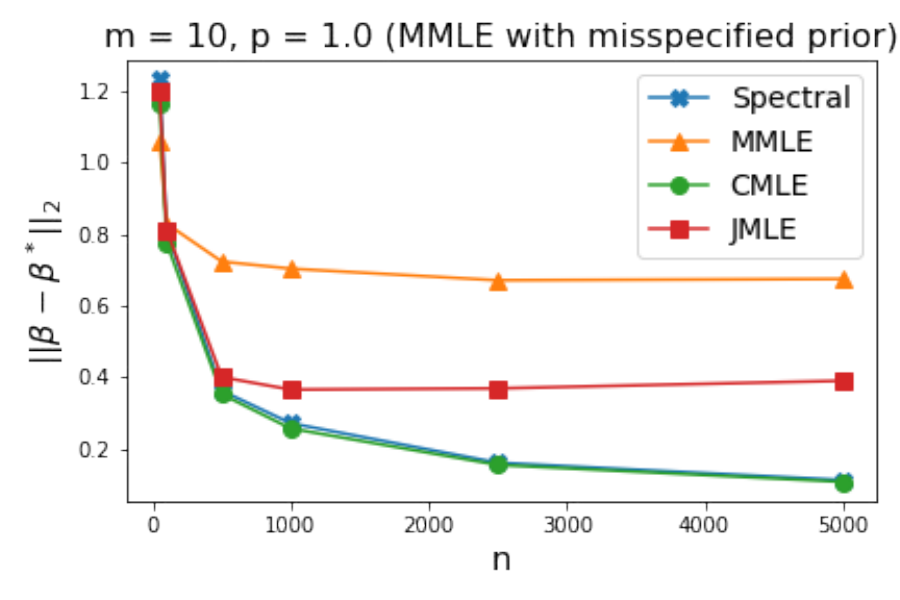}
  \caption{When the MMLE prior distribution is misspecified, its performance could degrade.}
  \label{fig:sfig2}
\end{subfigure}\\
\begin{subfigure}{.45\textwidth}
  \centering
  \includegraphics[height=40mm, width=60mm]{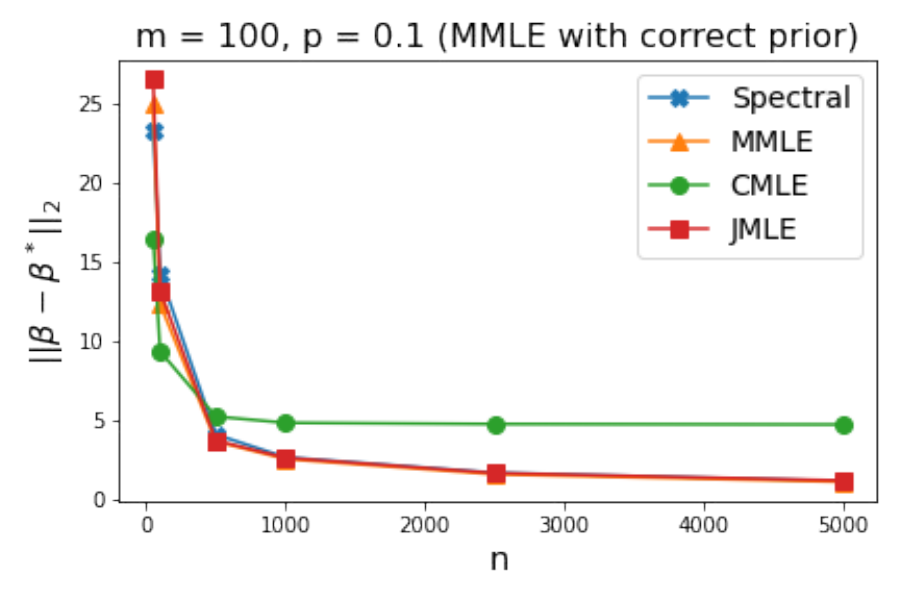}
  \caption{In larger datasets with missing data, CMLE's performance could degrade.}
  \label{fig:sfig3}
\end{subfigure}
\begin{subfigure}{.45\textwidth}
  \centering
  \includegraphics[height=40mm, width=60mm]{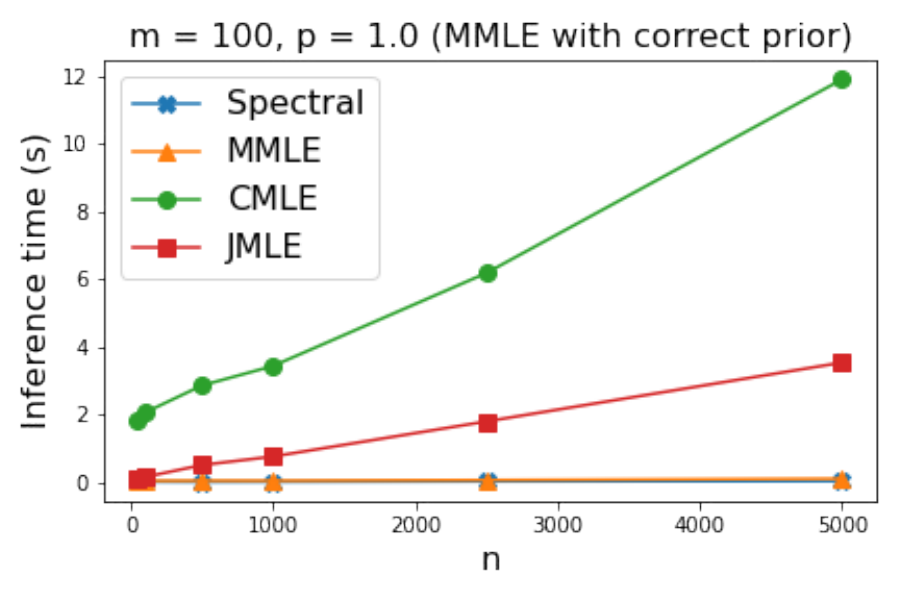}
  \caption{The spectral method is comparatively scalable.}
  \label{fig:sfig4}
\end{subfigure}
\caption{\emph{(Synthetic Data Experiments.)} The spectral method performs consistently well (both in terms of $\ell_2$ error and time complexity) across a range of settings while CMLE, MMLE and JMLE could underperform in unfavorable settings. Presented results have been averaged over 100 trials.}
\label{fig:synthetic}
\end{figure}

\textbf{Synthetic Data:} We generate the item parameters $\beta^*$ from a standard normal distribution and user parameters $\theta^*$ from $\calN(0, \sigma^2)$ where $\sigma^2$ varies for different model settings. Recall that MMLE requires the statistician to specify the prior distribution over the user parameters and for synthetic experiments, we specify this prior distribution to be the standard normal distribution. Subfigures (a)-(c) of Figure \ref{fig:synthetic} show $\lVert \beta -\beta^*\rVert_2$ against $n$ for all 4 algorithms under 3 \emph{different model settings} while Subfigure (d) shows inference time. When there is no missing data and MMLE's prior distribution is correctly specified ($\sigma = 1$), CMLE, MMLE and the spectral algorithm perform equally well. However, when MMLE's prior distribution is misspecified ($\sigma = 2$), it produces inconsistent estimates. As mentioned before, JMLE is known to produce inconsistent estimate when $m$ is small relative to $n$. On the other hand, CMLE's performance degrades under moderately sized dataset with missing data. The spectral algorithm, however, consistently performs well across all of these settings and is comparatively scalable.

\textbf{Real Data:} We perform experiments on a wide range of real-life datasets from education testing datasets to book and movie ratings datasets for recommendation systems. In order to transform ratings data to binary response data, we follow the procedures in previous works \cite{lan2018estimation, davenport20141}. Specifically, for each user, we convert all ratings higher than the average to 0 and 1 otherwise (so that items with a higher parameter value is `better'). Since real-life datasets do not come with true $\beta^*$ parameters, we compare the algorithms on four metrics: area under the ROC curve (denoted AUC) on heldout test data; log-likelihood on heldout test data; inference time and top-$K$ accuracy where the reference top-$K$ set is determined by average ratings. We have also followed the standard procedure in the recommendation systems literature to remove items that have few ($\leq 10$) ratings from the reference top-$K$ set. In our experiments, we consider $K \in \{ 10, 25, 50\}$. We defer extra experiment results where we include additional algorithms such as Bayesian estimation \cite{natesan2016bayesian} and pairwise likelihood estimation to the supplementary materials. We mention here a few notable datasets: RIIID \cite{riiid} ($m=6k, n=23k$, education testing dataset), ML-20M \cite{harper2015movielens} ($m=27k,n=138k$), Book-Genome \cite{kotkov2022tag} ($m=10k, n=350k$).

Table \ref{tbl:all-results-main} summarizes the performance of the four methods in our experiments. Note that the first four datasets are education testing datasets and thus there are no top-$K$ ranking metrics being measured. For tuning the prior distribution for MMLE, we select the prior distribution that admits the highest log-likelihood on a validation set. On the other hand, CMLE, JMLE and the spectral method requires minimal model tuning. For small scale education datasets (LSAT, UCI, 3GRADES), there are no missing responses and all methods perform very similarly to one another. For large scale ratings datasets, the spectral method remains competitive with the baselines in terms of AUC and log-likelihood but tends to outperforms in top-$K$ accuracy and is significantly more efficient.

In large scale ratings datasets, we observe that the competitor methods tend to assign large parameter value to items that receive only a few but very high ratings (so after data processing, all of the responses to such items are $0$) and these items are not included in the reference top-$K$ set. The spectral method, because it operates on pairwise differentials, is less susceptible to these noisy responses and thus more accurately recovers the items in the reference top-$K$ set.

\begin{table}[]
\begin{adjustwidth}{}{}
\resizebox{1.0\textwidth}{!}{%
\begin{tabular}{|c|cccc|cccc|cccc|cccc|}
\hline
          & \multicolumn{4}{c|}{\large AUC}  & \multicolumn{4}{c|}{\large Log-likelihood}   & \multicolumn{4}{c|}{\large Top-$[10\git 25 \git 50]$ accuracy}  & \multicolumn{4}{c|}{\large Total Inference time (seconds)}  \\ \hline
\large Dataset   & \multicolumn{1}{c|}{\large Spectral} & \multicolumn{1}{c|}{ MMLE} & \multicolumn{1}{c|}{CMLE} & JMLE & \multicolumn{1}{c|}{\large Spectral} & \multicolumn{1}{c|}{MMLE} & \multicolumn{1}{c|}{CMLE} & JMLE & \multicolumn{1}{c|}{\large Spectral} & \multicolumn{1}{c|}{MMLE} & \multicolumn{1}{c|}{CMLE} & JMLE & \multicolumn{1}{c|}{\large Spectral} & \multicolumn{1}{c|}{MMLE} & \multicolumn{1}{c|}{CMLE} & JMLE \\ \hline
 {LSAT} & \multicolumn{1}{c|}{$0.707$} & \multicolumn{1}{c|}{$0.707$} & \multicolumn{1}{c|}{$0.707$}  & {$0.707$} & \multicolumn{1}{c|}{$-0.487$} & \multicolumn{1}{c|}{$-0.489$} & \multicolumn{1}{c|}{$-0.487$} & {\shade $-0.485$}  & \multicolumn{1}{c|}{N/A} & \multicolumn{1}{c|}{N/A} & \multicolumn{1}{c|}{N/A} & {N/A} & \multicolumn{1}{c|}{\shade $0.028$} & \multicolumn{1}{c|}{$0.159$} & \multicolumn{1}{c|}{$0.154$} & {$0.075$} \\ \hline
 {UCI} & \multicolumn{1}{c|}{$0.565$} & \multicolumn{1}{c|}{$0.565$} & \multicolumn{1}{c|}{$0.565$}  & {$0.565$} & \multicolumn{1}{c|}{$-0.687$} & \multicolumn{1}{c|}{\shade $-0.686$} & \multicolumn{1}{c|}{$-0.692$} & {$-0.706$}  & \multicolumn{1}{c|}{N/A} & \multicolumn{1}{c|}{N/A} & \multicolumn{1}{c|}{N/A} & {N/A} & \multicolumn{1}{c|}{\shade $0.015$} & \multicolumn{1}{c|}{$0.133$} & \multicolumn{1}{c|}{$0.136$} & {$0.034$} \\ \hline
 {3 GRADES} & \multicolumn{1}{c|}{$0.532$} & \multicolumn{1}{c|}{$0.532$} & \multicolumn{1}{c|}{$0.532$}  & {$0.532$} & \multicolumn{1}{c|}{$-0.706$} & \multicolumn{1}{c|}{\shade $-0.692$} & \multicolumn{1}{c|}{$-0.699$} & {$-0.717$}  & \multicolumn{1}{c|}{N/A} & \multicolumn{1}{c|}{N/A} & \multicolumn{1}{c|}{N/A} & {N/A} & \multicolumn{1}{c|}{$0.021$} & \multicolumn{1}{c|}{$0.181$} & \multicolumn{1}{c|}{$0.105$} & {\shade $0.009$} \\ \hline
 {RIIID} & \multicolumn{1}{c|}{$0.723$} & \multicolumn{1}{c|}{\shade $0.724$} & \multicolumn{1}{c|}{N/A}  & {$0.724$} & \multicolumn{1}{c|}{\shade $-0.486$} & \multicolumn{1}{c|}{$-0.49$} & \multicolumn{1}{c|}{N/A} & {$-0.486$}  & \multicolumn{1}{c|}{N/A} & \multicolumn{1}{c|}{N/A} & \multicolumn{1}{c|}{N/A} & {N/A} & \multicolumn{1}{c|}{\shade $13.1$} & \multicolumn{1}{c|}{$104$} & \multicolumn{1}{c|}{N/A} & {$61.2$} \\ \hline
 {HETREC} & \multicolumn{1}{c|}{\shade $0.729$} & \multicolumn{1}{c|}{$0.729$} & \multicolumn{1}{c|}{$0.506$}  & {$0.73$} & \multicolumn{1}{c|}{$-0.604$} & \multicolumn{1}{c|}{$-0.603$} & \multicolumn{1}{c|}{$-1.119$} & {\shade $-0.602$}  & \multicolumn{1}{c|}{\shade $0.5 \git 0.64 \git 0.6$} & \multicolumn{1}{c|}{$0.0 \git 0.0 \git 0.02$} & \multicolumn{1}{c|}{$0.0 \git 0.0 \git 0.0$} & {$0.0 \git 0.0 \git 0.02$} & \multicolumn{1}{c|}{\shade $50.1$} & \multicolumn{1}{c|}{$140$} & \multicolumn{1}{c|}{$224 \text{k}$} & {$144$} \\ \hline
 {ML-100K} & \multicolumn{1}{c|}{$0.662$} & \multicolumn{1}{c|}{$0.659$} & \multicolumn{1}{c|}{$0.498$}  & {\shade $0.665$} & \multicolumn{1}{c|}{\shade $-0.646$} & \multicolumn{1}{c|}{$-0.66$} & \multicolumn{1}{c|}{$-1.159$} & {$-0.653$}  & \multicolumn{1}{c|}{\shade $0.4 \git 0.6 \git 0.54$} & \multicolumn{1}{c|}{$0.0 \git 0.0 \git 0.0$} & \multicolumn{1}{c|}{$0.0 \git 0.0 \git 0.0$} & {$0.0 \git 0.0 \git 0.0$} & \multicolumn{1}{c|}{\shade $1.39$} & \multicolumn{1}{c|}{$16.2$} & \multicolumn{1}{c|}{$9.56 \text{k}$} & {$21$} \\ \hline
 {ML-1M} & \multicolumn{1}{c|}{$0.698$} & \multicolumn{1}{c|}{\shade $0.701$} & \multicolumn{1}{c|}{$0.468$}  & {$0.7$} & \multicolumn{1}{c|}{\shade $-0.626$} & \multicolumn{1}{c|}{$-0.632$} & \multicolumn{1}{c|}{$-1.166$} & {$-0.63$}  & \multicolumn{1}{c|}{\shade $0.8 \git 0.72 \git 0.72$} & \multicolumn{1}{c|}{$0.6 \git 0.6 \git 0.62$} & \multicolumn{1}{c|}{$0.0 \git 0.0 \git 0.0$} & {$0.5 \git 0.64 \git 0.66$} & \multicolumn{1}{c|}{\shade $19.2$} & \multicolumn{1}{c|}{$86.9$} & \multicolumn{1}{c|}{$156 \text{k}$} & {$194$} \\ \hline
 {EACH MOVIE} & \multicolumn{1}{c|}{$0.716$} & \multicolumn{1}{c|}{\shade $0.718$} & \multicolumn{1}{c|}{$0.522$}  & {$0.716$} & \multicolumn{1}{c|}{$-0.615$} & \multicolumn{1}{c|}{\shade $-0.613$} & \multicolumn{1}{c|}{$-0.946$} & {$-0.614$}  & \multicolumn{1}{c|}{\shade $0.8 \git 0.76 \git 0.82$} & \multicolumn{1}{c|}{$0.8 \git 0.68 \git 0.84$} & \multicolumn{1}{c|}{$0.0 \git 0.0 \git 0.02$} & {$0.6 \git 0.6 \git 0.72$} & \multicolumn{1}{c|}{\shade $11.3$} & \multicolumn{1}{c|}{$329$} & \multicolumn{1}{c|}{$220 \text{k}$} & {$1.9 \text{k}$} \\ \hline
 {ML-10M} & \multicolumn{1}{c|}{$0.714$} & \multicolumn{1}{c|}{\shade $0.716$} & \multicolumn{1}{c|}{N/A}  & {$0.716$} & \multicolumn{1}{c|}{\shade $-0.617$} & \multicolumn{1}{c|}{$-0.619$} & \multicolumn{1}{c|}{N/A} & {$-0.618$}  & \multicolumn{1}{c|}{\shade{$0.5 \git 0.84 \git 0.7$} }  & \multicolumn{1}{c|}{$0.1 \git 0.28 \git 0.32$} & \multicolumn{1}{c|}{N/A} & {$0.0 \git 0.32 \git 0.36$} & \multicolumn{1}{c|}{\shade $821$} & \multicolumn{1}{c|}{$3.93 \text{k}$} & \multicolumn{1}{c|}{N/A} & {$6.55 \text{k}$} \\ \hline
 {ML-20M} & \multicolumn{1}{c|}{\shade $0.72$} & \multicolumn{1}{c|}{$0.71$} & \multicolumn{1}{c|}{N/A}  & { $0.71$} & \multicolumn{1}{c|}{\shade $-0.619$} & \multicolumn{1}{c|}{$-0.619$} & \multicolumn{1}{c|}{N/A} & {$-0.619$}  & \multicolumn{1}{c|}{\shade $0.5 \git 0.8 \git 0.64$} & \multicolumn{1}{c|}{$0.3 \git 0.44 \git 0.4$} & \multicolumn{1}{c|}{N/A} & {$0.1 \git 0.4 \git 0.4$} & \multicolumn{1}{c|}{\shade $1.58 \text{k}$} & \multicolumn{1}{c|}{$5.36 \text{k}$} & \multicolumn{1}{c|}{N/A} & {$4.42 \text{k}$} \\ \hline
 {BX} & \multicolumn{1}{c|}{$0.546$} & \multicolumn{1}{c|}{\shade $0.577$} & \multicolumn{1}{c|}{$0.503$}  & {$0.57$} & \multicolumn{1}{c|}{$-0.618$} & \multicolumn{1}{c|}{\shade $-0.612$} & \multicolumn{1}{c|}{$-0.8$} & {$-0.617$}  & \multicolumn{1}{c|}{$0.3 \git 0.16 \git 0.16$} & \multicolumn{1}{c|}{\shade $0.3 \git 0.24 \git 0.2$} & \multicolumn{1}{c|}{$0.0 \git 0.0 \git 0.02$} & {$0.3 \git 0.2 \git 0.18$} & \multicolumn{1}{c|}{\shade $205$} & \multicolumn{1}{c|}{$2.02 \text{k}$} & \multicolumn{1}{c|}{$156 \text{k}$} & {$481$} \\ \hline
 {BOOK-GENOME} & \multicolumn{1}{c|}{$0.658$} & \multicolumn{1}{c|}{\shade $0.665$} & \multicolumn{1}{c|}{N/A}  & {$0.654$} & \multicolumn{1}{c|}{$-0.651$} & \multicolumn{1}{c|}{\shade $-0.645$} & \multicolumn{1}{c|}{N/A} & {$-0.651$}  & \multicolumn{1}{c|}{\shade $0.6 \git 0.44 \git 0.42$} & \multicolumn{1}{c|}{$0.3 \git 0.32 \git 0.34$} & \multicolumn{1}{c|}{N/A} & {$0.2 \git 0.24 \git 0.38$} & \multicolumn{1}{c|}{\shade $2.53 \text{k}$} & \multicolumn{1}{c|}{$2.56 \text{k}$} & \multicolumn{1}{c|}{N/A} & {$4.34 \text{k}$} \\ \hline
\end{tabular}}
\end{adjustwidth}
\setlength{\belowcaptionskip}{-8pt}
\caption{\emph{(Real Data Experiments.)} The spectral method (1st column under each metric) is competitive with the baselines (best results are shaded) especially in terms of ranking metrics \protect\footnotemark. The spectral method is generally the fastest method on large datasets. It \emph{works off-the-shelf with minimal tuning} and is \emph{comparatively accurate}.\label{tbl:all-results-main}}
\end{table}

\footnotetext{As noted before, CMLE has been observed to underperform on large datasets with missing data. In some of our experiments, CMLE fails due to numerical errors or does not converge. The results for CMLE are marked `N/A' for these experiments.}

\section{Related Works}
The Rasch modeling literature is quite broad and we refer the interested reader to recent surveys \cite{aryadoust2021comprehensive,robitzsch2021comprehensive}. Since its original formulation to model psychological tests outcome \cite{rasch1960studies}, the model has been extended to account for more complicated response patterns such as numerical ratings and ordinal responses \cite{andersen1977sufficient,andrich1978rating,wright1982rating,masters1982rasch}. Higher-order models that incorporates bias variables such as the 2PL and 3PL model \cite{birnbaum1968some} and multivariate models \cite{embretson1984general} remain active areas of research where machine learning techniques have recently been applied with substantial success \cite{bergner2012model}. 
The Rasch inference problem is also closely connected to the 1-bit matrix completion problem in machine learning where we observe a sparse $n\times m$ binary matrix with underlying entrywise probability $f(M)$ where $f$ is a mapping function (e.g., logistic) and $M$ is a real-valued matrix. There, the goal is obtain an estimate of $M$. A commonly proposed approaches for 1-bit matrix completion based on alternating optimization is exactly joint maximum likelihood estimate \cite{chen2019joint,chen2021note}.

% In recent years, there has been 
% A recent publication \cite{lan2018estimation} proposed an estimation framework for a modification of the Rasch model using the probit link function as opposed to the logistic link function in Equation (\ref{eqn:rasch-prob}). There the goal is to establish an analysis framework that provides finite sample error bounds and the authors proposed a \emph{joint parameter} estimation method based on probit regression.
% As noted before, our algorithm is also related to the Rank Centrality algorithm \cite{negahban2017rank} for parameter estimation under the Bradley-Terry-Luce model \cite{bradley1952rank,luce2012individual} and often referred to as spectral ranking \cite{maystre2015fast,agarwal2018accelerated}. Broadly speaking, both Rank Centrality and our algorithm fall within the class of spectral algorithms of which the central idea is the computation of the leading eigenvector of a matrix. The spectral approach is thus a general recipe and we apply this approach to the Rasch model estimation problem. 

\section{Ethical Considerations}

Our work proposes an algorithm of which real-life applications very often involve actual human data with sensitive information. For example, the Rasch model is often studied in the context of education testing and psychological testing where the subjects of studies are students and patients. Therefore, deploying our algorithm (or any algorithms in this context) needs to be accompanied by thoughtful and thorough ethical considerations. In this work, we provide the algorithmic tool that lays the foundation for our algorithm and its theoretical guarantee. We believe that a socially constructive application of our algorithm should always be accompanied by detailed explanation of its fundamental limitations, assumptions and decision makers need to take into account these aspects when interpreting the results returned by the algorithm.

% \vspace{-0.5em}
\section{Conclusion}
% \vspace{-0.5em}
We propose a new spectral algorithm for the item estimation problem under the celebrated Rasch model. Our algorithm is theoretically well-founded, practically performant and should be added to the statistician's quiver of estimation methods when analyzing binary response data. Extending our algorithm to more expressive IRT models such as 2PL or 3PL and response types is an open avenue. In the future, we also hope to generalize the method to more complicated response data types such as ordinal or rating data, as well as incorporating ancillary information (user and item features).
% Another important aspect inline with the ethical considerations that need to go into IRT models is privacy. Given the human centric nature of IRT models and how IRT algorithms are often deployed on datasets containing sensitive information (e.g., a student passing or failing a test, a voter supporting a legislation), it is quite concerning that there has not been any proposed privacy-preserving estimation method in the literature. 

\section{Acknowledgement}
The authors would like to thank William Zhang for proofreading and leaving helpful comments on earlier drafts of this paper. We would also like to thank the anonymous reviewers for their thoughtful suggestions that have been incorporated to improve this paper. D.N. and A.Z. are supported by NSF Grant DMS-2112988. A.Z. acknowledges financial support from the Alfred H. Williams Faculty Scholar award. Any opinions expressed in this paper are those of the author and do not necessarily reflect the views of the National Science Foundation.

\bibliography{reference}

\newpage
\appendix

% \section{Appendix}
\section{Proofs of Upper Bounds}
\textbf{Proofs overview: }The starting point for our proof is a Markov chain eigen-perturbation bound- Lemma \ref{lem:perturb-bound-mc}. Without going too deeply into the details and definitions, we have
$$ \lVert \pi - \pi^* \rVert_2 \leq \frac{\lVert {\pi^*}^\top(P^* - P)\rVert_{2}  }{\mu^*(P^*) - \lVert P - P^* \rVert_{2} }\,,$$
where $\mu^*(P^*)$ is the spectral gap of the idealized Markov chain. We then bound the numerator and the denominator separately. We will show (see Corollary \ref{coro:spectral-error-difference}):
$$ \mu^*(P^*) - \lVert P - P^* \rVert_2 \gtrsim \frac{1}{d}\,, $$
where $\gtrsim$ denotes $\Omega$-asymptotic relation. For the regime where $m$ is a constant or grows very slowly with respect to $n$, we show (see Lemma \ref{lem:projected-error-bound}) that
$$ \lVert {\pi^*}^\top(P^* - P)\rVert_{2}  \lesssim \frac{\sqrt{\max \{m, \log np^2 \} }}{dm\sqrt{np^2}} \,, $$
where $\lesssim$ denote $O$-asymptotic relation. For the regime where $m$ grows ($mp \gtrsim \log n$), we show (see Lemma \ref{lem:projected-error-bound-growing-m}) that
$$ \lVert {\pi^*}^\top(P^* - P)\rVert_{2}  \lesssim \frac{1}{d\sqrt{mnp}} \,. $$
Lastly, we will show that
$$ \lVert \beta - \beta^* \rVert_2 \lesssim m \cdot \lVert \pi- \pi^*\rVert_2\,. $$

\subsection{Preliminaries} 
Recall that $\pi^*_i = e^{\beta^*_i}/\left(\sum_{j=1}^m \beta^*_j\right)$ for $i \in [m]$. Define $\pi^*_{\max} := \max_{i\in [m]} \pi^*_i$ and $\pi^*_{\min} := \min_{i\in [m]} \pi_i^*$. Define $ \kappa := \beta^*_{\max} - \beta^*_{\min}$, then $\pi^*_{\max}/\pi^*_{\min} \leq e^{\kappa}$. Let $\gamma := \min_{l\in[n], i, j\in [m]} \E[X_{li}(1-X_{lj})]$.

Define $B = A^\top A$ and the following events:
$$ \A = \{ \frac{np^2}{2} \leq B_{ij} \leq \frac{3np^2}{2} \,\forall i\neq j\in [m] \}\,,$$
$$ \A^+ = \A \cap \{ \frac{mp}{2} \leq A_{l}^\top \mb 1 \leq \frac{3mp}{2} \,\forall l \in [n] \} \,.$$
Both events happen with high probability under appropriate conditions, as summarized by two lemmas below.

\begin{lemma} Consider the random sampling scheme described in Section \ref{sect:error-guarantees}, we have
$$ \Pr(\A) \geq 1 -  \exp{-\frac{np^2}{20}} $$
so long as $np^2 \geq C_1 \log m$ by a sufficiently large constant $C_1$ (e.g., $C_1 \geq 60$).
\end{lemma}
\begin{proof} Invoking Chernoff bound, we have:
\begin{equation*}
\begin{aligned}
\Pr(\lvert \sum_{l=1}^n A_{li}A_{lj} - \E[A_{li}A_{lj}] \rvert > \frac{1}{2}np^2) \leq 2\exp{-\frac{\frac{1}{4}np^2 }{\frac{1}{2}+2}}
= \exp{-\frac{np^2}{10}+\ln 2}\,.
\end{aligned}
\end{equation*}
It can be checked that so long as $np^2 \geq 60\ln m \geq 20\ln 2 + 40\ln m$ for $m \geq 2$, then $\exp{-\frac{np^2}{10}+\ln 2} \leq \exp{-\frac{np^2}{20} -2\ln m } $. The rest of the proof follows by applying union bound over all pairs $i\neq j$.
\end{proof}

\begin{lemma} Consider the random sampling scheme described in Section \ref{sect:error-guarantees}, we have
$$ \Pr(\A^+) \geq 1 - \exp{-\frac{np^2}{20}} - \frac{1}{n^{9}} $$
so long as $np^2 \geq C_1 \log m$ and $mp \geq C_1 \log n$ by a sufficiently large constant $C_1$ (e.g., $C_1 > 101$).
\end{lemma}
\begin{proof} The first term $\exp{-\frac{np^2}{20}}$ is obtained using the same argument as in the lemma above. For the second term $\frac{1}{m^9}$, we again invoke Chernoff bound:
\begin{equation*}
\begin{aligned}
\Pr(\lvert \sum_{l=1}^n A_{li}A_{lj} - \E[A_{li}A_{lj}] \rvert > \frac{1}{2}mp) \leq 2\exp{-\frac{\frac{1}{4}mp }{\frac{1}{2}+2}}
= \exp{-\frac{mp}{10}+\ln 2}\,.
\end{aligned}
\end{equation*}
So long as $mp \geq 100 \log n$, one could see that $\exp{-\frac{mp}{10}+\ln 2} \leq \frac{1}{n^{10}}$. Applying union bound over all users $n$ gives us the probability bound.
\end{proof}

For the rest of the proof, we assume that either event $\A$ (or $\A^+$) happens and simply set $d = \frac{3mnp^2}{2} $. One can rigorously justify that any valid choice of the normalization factor $d$ so long as no entries of the Markov chain is negative will yield the same final output (modulo a transformation). This is an application of Theorem \ref{thm:equivalent-markov-chains}. We also state without proof the following useful identities.
$$ \pi^*_{\min} \geq \frac{1}{me^{\kappa}}\,. $$
$$ \pi^*_{\max} \leq \frac{e^{\kappa}}{m}\,. $$

~\\
\textbf{Reversibility of the idealized Markov chain:} Note that the eigenperturbation bound in Lemma \ref{lem:perturb-bound-mc} requires that the reference Markov chain $P^*$ is reversible. Fortunately for us, this is indeed the case.

\begin{repproposition}{lem:reversibility} For a fixed assignment $A$, consider the following idealized the Markov chain.
$$ P^*_{ij} =\begin{cases} \frac{1}{d} \sum_{l=1}^n A_{li}A_{lj} \E[X_{li}(1-X_{lj})] &\text{for } i\neq j \\ 1 - \frac{1}{d} \sum_{k\neq i} P^*_{ik} &\text{for } i = j \end{cases}\quad , $$
where $d$ is some sufficiently large normalization constant. Then the Markov chain $P^*$ is reversible and its stationary distribution is $\pi^*$. 
\end{repproposition}

\begin{proof} This boils down to verifying the reversility condition, i.e., whether $\pi_i^* P_{ij}^* = \pi_j^* P_{ji}^*$. One can see that
\begin{equation*}
\begin{aligned}
\pi_i^* P^*_{ij} &= e^{\beta_i^*} \cdot \frac{1}{\sum_{k\in[m]} e^{\beta_k^*} }\cdot \sum_{l\in [m]} A_{li}A_{lj} \frac{e^{\theta^*_l}}{e^{\theta^*_l} + {e^{\beta^*}}_i} \cdot \frac{e^{\beta_j^*}}{e^{\theta^*_l} + e^{\beta_j^*}} \\
&= e^{\beta_j^*} \cdot \frac{1}{\sum_{k\in[m]} e^{\beta_k^*}}\cdot\sum_{l\in [m]}A_{li}A_{lj} \frac{e^{\theta^*_l}}{e^{\theta^*_l} + {e^{\beta^*}}_j} \cdot \frac{e^{\beta_i^*}}{e^{\theta^*_l} + e^{\beta_i^*}} \\
&= \pi_j^* P^*_{ji}\,.
\end{aligned}
\end{equation*}
This completes the proof.
\end{proof}

~\\
\textbf{$\ell_2$ eigen-perturbation bound:} The reader might recognize the similarity between the following perturbation bound to Theorem 8 of \cite{chen2019spectral}. Our lemma has been modified to use the $\ell_2$ norm instead of the induced norm in the original theorem.

\begin{lemma}\label{lem:perturb-bound-mc} Consider two discrete time Markov chains $P$ and $P^*$ with a finite state space and stationary distributions $\pi$ and $\pi^*$, respectively. Furthermore, assume that the Markov chain $P^*$ is reversible. If $\lVert P - P^* \rVert_2\leq \mu^*(P^*)$ where $\mu^*(P^*)$ is the spectral gap of $P^*$, then
$$ \lVert \pi - \pi^* \rVert_{2} \leq \frac{\lVert {\pi^*}^\top(P^* - P)\rVert_{2}  }{\mu^*(P^*) - \lVert P - P^* \rVert_{2} }\,. $$
\end{lemma}

\begin{proof} We have:
\begin{equation*}
\begin{aligned}
{\pi^*}^\top - \pi^\top &= {\pi^*}^\top P^* - \pi^\top P\\
&=  {\pi^*}^\top (P^* - P + P) - \pi^\top P \\
&= {\pi^*}^\top (P^* - P) + {\pi^*}^\top P - \pi^\top P \\
&=  {\pi^*}^\top (P^* - P) + {(\pi^* - \pi)}^\top P\\
&= {\pi^*}^\top (P^* - P) + {(\pi^* - \pi)}^\top (P- P^* + P^*)\\
&= {\pi^*}^\top (P^* - P) + {(\pi^* - \pi)}^\top (P- P^*) +  ({\pi^* - \pi})^\top P^*\\
&= {\pi^*}^\top (P^* - P) + {(\pi^* - \pi)}^\top (P- P^*) +  ({\pi^* - \pi})^\top (P^* - \mb 1 {\pi^*}^\top )\,.\\
\end{aligned}
\end{equation*}
The last equality comes from the simple observation that $(\pi^* - \pi)^\top \mb 1 = 0$. We thus obtain the following normed inequality:
$$ \lVert \pi - \pi^* \rVert_2 \leq \lVert{\pi^*}^\top (P^* - P)\rVert_2 + \lVert{\pi^* - \pi}\lVert_2 \cdot \rVert P- P^*\rVert_2 +  \lVert{\pi^* - \pi}\rVert_2 \cdot \rVert P^* - \mb 1 {\pi^*}^\top\rVert_2 \,.$$
If we can show $1-\rVert P^* - \mb 1 {\pi^*}^\top\rVert_2$ is exactly the spectral gap of the reversible Markov chain $P^*$, then the final inequality is readily obtained after a simple rearrangment. We devote the rest of the proof towards this.

Because $P^*$ is reversible, i.e., $\pi^*_i P^*_{ij} = \pi_j^* P_{ji}^*$, it can be checked that the matrix $\Lambda^{1/2} P^* \Lambda^{-1/2}$ is symmetric and so is $\Lambda^{1/2} \mb 1{\pi^*}^\top \Lambda^{-1/2}$. Because there is a similarity transformation between $P^* - \mb 1{\pi^*}^\top$ and $\Lambda^{1/2} (P^*- \mb 1{\pi^*}^\top) \Lambda^{-1/2}$, it suffices to analyze the spectrum of the later symmetric matrix. Let $v = [\sqrt \pi_1, \ldots, \sqrt \pi_m]$ (it has unit length). It can be checked that
\begin{enumerate}
\item $v^\top \Lambda^{1/2}P^*\Lambda^{-1/2} = \pi^\top P^* \Lambda^{-1/2} = v^\top$. Essentially, $v$ is a eigenvector associated with eigenvalue 1.
\item $\Lambda^{1/2}\mb 1 {\pi^*}^\top \Lambda^{-1/2} = 1 vv^\top$.
\end{enumerate}
These two observations and the elementary fact that a Markov chain has leading eigenvalue 1 readily imply that $1- \lVert P^* - \mb 1 {\pi^*}^\top \rVert_2$ is exactly the spectral gap of $P^*$.
\end{proof}

\subsection{Bounding the spectral gap}
Towards bounding the denominator of the eigen-perturbation bound, we will firsrt lower bound the spectral gap of the idealized Markov chain $P^*$. We first state the following useful comparison lemma that has been presented as Lemma 2 of \cite{negahban2017rank} and is originally due to \cite{diaconis1993comparison}.

\begin{lemma}\label{lem:comparison-lemma} Consider two reversible Markov chains $P^*, Q$ with stationary distribution $\pi^*, \pi$, respectively that are defined on the same graph $G(V, E)$ of $m$ states. That is, $P^*_{ij} = 0$ and $Q_{ij} =0$ if $i, j\notin E$. Define $\alpha = \min_{i, j\in E} \frac{\pi^*_i P{ij}^* }{\pi_i Q_{ij} }  $ and $\beta = \max_{i} \frac{\pi_i^*}{\pi_i}$. We have
$$ \frac{\mu^*(P^*) }{\mu^*(Q)} \geq \frac{\alpha}{\beta}\,, $$
where $\mu^*(.)$ is the spectral gap operator.
\end{lemma}

Note that the comparison lemma lower bounds the spectral gap of a reversible Markov chain in terms of another reversible Markov chain. 
Considera Markov chain whose pairwise transition probabilties are defined as follows:
\begin{equation}\label{eqn:reference-mc}
Q_{ij} = \begin{cases} \frac{B_{ij}}{d} &\text{for } i\neq j \\ 1 - \frac{1}{d} \sum_{k\neq i} B_{ik} &\text{for } i= j  \end{cases}
\end{equation}
where $d$ is the same normalization constant as in the idealized Markov chain described in Lemma \ref{lem:reversibility}.
By design, this is a reversible Markov chain whose stationary distribution is the uniform distribution, $q = \frac{1}{m} \mb 1_m$.

\begin{lemma}\label{lem:spectral-gap-reference} Conditioned on event $\A$,
$$ \mu^*(Q) \geq \frac{1}{3}\,. $$
\end{lemma}

\begin{proof} Let $\lambda_{\max, \indep}(Q)$ denote the second largest eigenvalue of $Q$ and $D = \diag(\mb 1/d)$. We have:
\begin{equation*}
\begin{aligned}
Q&= I  - D^{-1} \diag(B^\top \mb 1) + D^{-1}B\\
\Rightarrow \lambda_{\max, \indep} (Q) &= \lambda_{\max, \indep} (I  - D^{-1} \diag(B^\top \mb 1) + D^{-1}B)\\
 &= \lambda_{\max, \indep} (I  - \underbrace{\big[D^{-1} \diag(B^\top \mb 1) - D^{-1}B\big]}_{Laplacian} )\\
&= 1 - \lambda_{\min, \indep} (D^{-1} \diag(B^\top \mb 1) - D^{-1}B)\\
\Rightarrow 1 - \lambda_{\max, \indep}(Q) &= \lambda_{\min, \indep} (D^{-1} \diag(B^\top \mb 1) - D^{-1}B)
\end{aligned}
\end{equation*}
In these derivation steps, we have made use of the fundamental property of the Laplacian of a weighted graph: it has an eigenvalue 0 corresponding to the eigenvector proportional to $\mb 1_m$.
We now need to lower bound $\lambda_{\min, \indep} (D^{-1} \diag(B^\top \mb 1) - D^{-1}B)$. Conditioned on event $\A$:
\begin{equation*}
\begin{aligned}
\lambda_{\min, \indep} (D^{-1} \diag(B^\top \mb 1) - D^{-1}B) &= \frac{1}{d} \cdot \lambda_{\min, \indep}( B^\top \mb 1 -B)\\
&=  \frac{1}{d}\cdot\min_{u \indep \mb 1, \lVert u \rVert_2 = 1} \sum_{ij} (u_i - u_j)^2 B_{ij} \\
&\geq  \frac{1}{d}\cdot\frac{1}{2}np^2 \cdot \min_{u \indep \mb 1_m, \lVert u \rVert_2 = 1} \sum_{ij} (u_i - u_j)^2\\
&\geq  \frac{1}{d}\cdot\frac{1}{2}np^2 \cdot \min_{u \indep \mb 1_m, \lVert u \rVert_2 = 1} \, u^\top \big[m \, I_m - \mb 1_m \mb 1_m^\top \big] u = \frac{1}{2d}mnp^2\,.
\end{aligned}
\end{equation*}
Substituting $d=\frac{3}{2}mnp^2$ completes the proof.
\end{proof}

\begin{lemma}\label{lem:spectral-gap} Conditioned on event $\A$,
$$\mu(P^*) \geq \frac{\gamma}{3e^{2\kappa}}\,, $$
where $\gamma = \min_{l\in[n], i, j \in [m]} \E[X_{li}(1-X_{lj})]$.
\end{lemma}

\begin{proof} To prove the above lower bound, we will combine Lemmas \ref{lem:comparison-lemma}, the definition of the reference Markov chain $Q$ in Equation (\ref{eqn:reference-mc}), with stationary distribution $q = \frac{\mb 1}{m}$, with a lower bound on $\alpha$ and an upper bound on $\beta$. We have the following lower bound on $\alpha$.
\begin{equation*}
\begin{aligned}
\alpha &= \min_{i, j} \frac{\pi^*_i P^*_{ij}}{q_i Q_{ij} }\\
&= \min_{i, j} \frac{\pi^*_i P^*_{ij}}{\frac{1}{m} \cdot \frac{1}{d} \sum_{l=1}^n A_{li} A_{lj} }
= \min_{i, j} \frac{\pi^*_i  \frac{1}{d} \sum_{l=1}^n A_{li}A_{lj} \E[X_{li}(1-X_{lj})]  }{\frac{1}{m} \cdot \frac{1}{d} \sum_{l=1}^n A_{li} A_{lj} }\\
&= \min_{i, j} \frac{\pi^*_i   \sum_{l=1}^n A_{li}A_{lj} \E[X_{li}(1-X_{lj})]  }{\frac{1}{m} \cdot  \sum_{l=1}^n A_{li} A_{lj} }
\geq \min_{i, j} \frac{\pi^*_{\min} \gamma B_{ij}  }{\frac{1}{m} \cdot B_{ij} } = \frac{\pi^*_{\min}\gamma }{\frac{1}{m}}\\
&\geq \frac{\gamma }{e^{\kappa}}\,.
\end{aligned}
\end{equation*}
The last inequality follows from $ \pi ^*_{\min} \geq \frac{1}{m e^\kappa}$ (stated earlier). On the other hand, we have the following upper bound on $\beta$.
\begin{equation*}
\begin{aligned}
\beta &= \max_{i} \frac{\pi_i^*}{q_i} \leq \frac{1}{ \frac{1}{m}} \cdot \frac{e^{\kappa}}{m} \leq  e^{\kappa} \,.
\end{aligned}
\end{equation*}
Combining the lower bound above with the upper bound on $\alpha$ obtained earlier, we have
$$ \mu^*(P) \geq \frac{\gamma }{e^{2\kappa} } \cdot \mu^*(Q)\,. $$
The rest of the proof follows from the conclusion of Lemma \ref{lem:spectral-gap-reference}.
\end{proof}

\subsection{Bounding the matrix error term $\lVert P - P^* \rVert_2$}

\begin{lemma}\label{lem:operator-norm-bound} Suppose event $\A$ holds. Fix a small constant $\epsilon < 1$. Suppose further that $np^2\geq \frac{C_2\log m}{\gamma^2\epsilon^2}$ for a sufficiently large constant $C_2$ (e.g., $C_2 \geq 30$). Then
$$\lVert P - P^* \rVert_2 \leq 2\epsilon\gamma $$
with probability at least $1-\exp{-\frac{\gamma^2\epsilon^2np^2}{10}}$ over the random responses of the users where $\gamma = \min_{l\in [n], i\neq j \in [m]} \E[X_{li}(1-X_{lj})]$.
\end{lemma}

\begin{proof} Fix a pair $i, j$, let $\mu_{ij} := \frac{1}{n} \sum_{l=1}^n A_{li}A_{lj} \E[X_{li}(1-X_{lj})]$ (note that $\mu_{ij} \leq 1$). Applying Chernoff's bound gives us
\begin{equation*}   
\begin{aligned}
% &\Pr(\lvert \sum_{l=1}^n A_{li}A_{lj} X_{li}(1-X_{lj}) - A_{li}A_{lj}\E[X_{li}(1-X_{lj})] \rvert > \epsilon \E[\sum_{l=1}^n A_{li}A_{lj}\E[X_{li}(1-X_{lj})] )\\
&\Pr(\lvert \sum_{l=1}^n A_{li}A_{lj} X_{li}(1-X_{lj}) - A_{li}A_{lj}\E[X_{li}(1-X_{lj})] \rvert > \epsilon\gamma B_{ij}\,\lvert \, \A) \\
&=\Pr(\lvert \sum_{l=1}^n A_{li}A_{lj} X_{li}(1-X_{lj}) - A_{li}A_{lj}\E[X_{li}(1-X_{lj})] \rvert > \frac{\epsilon\gamma}{\mu_{ij}}\cdot  \mu_{ij} B_{ij} \,\lvert \,\A) \\\
&\leq 2\exp{-\frac{\gamma^2\epsilon^2/\mu_{ij}^2 \cdot \mu B_{ij} }{\frac{1}{2}+2}}\\
&\leq 2\exp{-\frac{\gamma^2\epsilon^2 \cdot \frac{np^2}{2} }{\frac{1}{2}+2}}\\
&= \exp{-\frac{\gamma^2\epsilon^2np^2}{5}+\ln 2}\,.
\end{aligned}
\end{equation*}
One can see that so long as $np^2 \geq \frac{30\ln m}{\gamma^2\epsilon^2}$ (and noting that $30\ln m \geq 20\ln m + 20\ln 2 \,\forall m \geq 2$) then $\exp{-\frac{\gamma^2\epsilon^2np^2}{5}+\ln 2} = \exp{-2\cdot \frac{\gamma^2\epsilon^2np^2}{10} +\ln 2} \leq \exp{-\frac{\gamma^2\epsilon^2np^2}{10}-2\ln m} $.
Applying union bound over all pairs $i\neq j$, we have with probability at least $1-\exp{-\frac{\gamma^2\epsilon^2np^2}{10}}$, $\lvert P_{ij} - P^*_{ij}\rvert \leq \frac{1}{d} \gamma\epsilon B_{ij} \leq \frac{3\gamma\epsilon np^2 }{2d}$ for all pairs $i\neq j$. We then have
\begin{equation*}
\begin{aligned}
\lVert P - P^* \rVert_{2} &\leq \lVert \diag(P) I - \diag(P^*) I\rVert_{2} + \lVert [P - P^*]_{i\neq j}\rVert_{2} \\
&\leq \max_{i} \lvert P_{ii} - P_{ii} \rvert + \max_{u, v: \lVert u \rVert = \lVert v\rVert = 1 } \sum_{i\neq j} u_i (P_{ij} - P_{ij}^*)v_j\\
&\leq \max_{i} \lvert \sum_{j\neq i} P_{ij}-P_{ij}^* \rvert + \max_{i\neq j} \lvert P_{ij}-P^*_{ij}\rvert \cdot \sum_{i\neq j} \lvert u_i\rvert \lvert v_j\rvert\\
&\leq 2 m \cdot \max_{i\neq j} \lvert P_{ij} - P_{ij}^* \rvert \\
&\leq \frac{3\epsilon\gamma mnp^2}{d} =  2\epsilon\gamma\,.\\
\end{aligned}
\end{equation*}
The conclusion follows from $d = \frac{3mnp^2}{2}$.
\end{proof}

Before moving on to bounding the projected error term, we first note that the eigen-perturbation bound in Lemma \ref{lem:perturb-bound-mc} holds when $\mu^*(P^*) \geq \lVert P - P^* \rVert_2$. We solve for $\epsilon$ such that
$$ 2\epsilon\gamma = \frac{1}{2} \cdot \frac{\gamma}{3e^{2\kappa}} \Rightarrow  \epsilon = \frac{1}{12e^{2\kappa}} \,. $$
We summarize this condition with the following corollary.
\begin{coro}\label{coro:spectral-error-difference} Conditioned on event $\A$ and suppose that 
$$np^2 \geq \frac{C_2 12^2 e^{4\kappa}}{\gamma^2} \log m $$
for sufficiently large constant $C_2$ (e.g., $C_2 \geq 30$). Then
$$ \mu^*(P) - \lVert P - P^* \rVert_2 \geq \frac{\gamma}{6 e^{2\kappa}} $$
with probability at least $1-\exp{-\frac{\gamma^2np^2}{10 \cdot 12^2 \cdot e^{4\kappa}}}$ over the random responses of the users.
\end{coro}

\begin{proof} Substituting $\epsilon = \frac{1}{12e^{2\kappa}}$ into the statement of Lemma \ref{lem:operator-norm-bound} gives: so long as $np^2 \geq \frac{30\gamma^2}{\big(  \frac{1}{12e^{2\kappa}} \big)^2} \cdot \log m$ the $\lVert P - P^* \rVert_2 \leq \mu^*(P^*)/2$ with probability at least $1-\exp{-\frac{\gamma^2np^2}{10} \cdot\big(  \frac{1}{12e^{2\kappa}} \big)^2  } $.
\end{proof}

\subsection{Bounding the projected error term $\lVert {\pi^*}^\top (P - P^*)\rVert_2 $}

\begin{lemma}\label{lem:projected-error-bound} Conditioned on event $\A$,
$$ \lVert {\pi^*}^\top (P - P^*)\rVert_2  \leq \frac{2e^\kappa \sqrt{16 \cdot \max\{m, \log np^2  \} }}{m \sqrt{np^2}}\, $$ 
with probability at least $1-  \min\{\exp{-12m}, \frac{1}{(np^2)^{12}} \} $ over the random responses of the users.
\end{lemma}

\begin{proof} We follow a simlar argument as in the proof of Lemma 8.4 in \cite{chen2020partial} in turning the normed term into a linear term. For completeness we reproduce this argument here. We first have
\begin{equation*}
\begin{aligned}
\lVert {\pi^*}^\top (P - P^*)\rVert_2 &= \sqrt{\sum_{i=1}^n\bigg(\sum_{j=1}^n \pi_j^*(P_{ji}-P_{ji}^*) \bigg)^2}\\
&= \sqrt{\sum_{i=1}^n\bigg(\sum_{j\neq i} \pi_j^*(P_{ji}-P_{ji}^*) + \pi_i^*(P_{ii}-P_{ii}^*) \bigg)^2}\\
&= \sqrt{\sum_{i=1}^n\bigg(\sum_{j\neq i} \pi_j^*(P_{ji}-P_{ji}^*) + \pi_i^*(\sum_{j\neq i} P^*_{ij}-P_{ij}) \bigg)^2}\\
&=  \sqrt{\sum_{i=1}^n\bigg(\sum_{j\neq i} \pi_j^*(P_{ji}-P_{ji}^*) + \pi_i^* (P^*_{ij}-P_{ij}) \bigg)^2}\,. \\
\end{aligned}
\end{equation*}
Let $\B$ denote the unit norm ball in $\R^m$ and $\mathcal{V}$ denote a $1/2$-net of $\B$. That is, for every $u \in \B$, there exists $v\in \mathcal{V}$ such that $\lVert u - v\rVert_2 \leq \frac{1}{2}$. 
For any $u\in \B$ and any corresponding $v$, we have
\begin{equation*}
\begin{aligned}
&\sum_{i=1}^n u_i\bigg(\sum_{j\neq i} \pi_j^*(P_{ji}-P_{ji}^*) + \pi_i^* (P^*_{ij}-P_{ij}) \bigg)\\
&= \sum_{i=1}^n v_i\bigg(\sum_{j\neq i} \pi_j^*(P_{ji}-P_{ji}^*) + \pi_i^* (P^*_{ij}-P_{ij}) \bigg) + \sum_{i=1}^n (u_i-v_i)\bigg(\sum_{j\neq i} \pi_j^*(P_{ji}-P_{ji}^*) + \pi_i^* (P^*_{ij}-P_{ij}) \bigg)\\
&\leq \sum_{i=1}^n v_i\bigg(\sum_{j\neq i} \pi_j^*(P_{ji}-P_{ji}^*) + \pi_i^* (P^*_{ij}-P_{ij}) \bigg) + \frac{1}{2} \cdot \sqrt{\sum_{i=1}^n \bigg(\sum_{j\neq i} \pi_j^*(P_{ji}-P_{ji}^*) + \pi_i^* (P^*_{ij}-P_{ij}) \bigg)^2}\,. \\
\end{aligned}
\end{equation*}
Maximizing both sides of the above inequality with respect to $u$ and rearranging the terms gives
\begin{equation*}
\begin{aligned}
&\sqrt{\sum_{i=1}^n \bigg(\sum_{j\neq i} \pi_j^*(P_{ji}-P_{ji}^*) + \pi_i^* (P^*_{ij}-P_{ij}) \bigg)^2}\\
&\leq 2 \max_{v\in \mathcal{V}} \sum_{i=1}^n v_i\bigg(\sum_{j\neq i} \pi_j^*(P_{ji}-P_{ji}^*) + \pi_i^* (P^*_{ij}-P_{ij}) \bigg)\,.
\end{aligned}
\end{equation*}
In summary, we can upper bound the normed term by a more manageable linear term as follows:
\begin{equation}\label{eqn:norm-linear}
\begin{aligned}
\lVert {\pi^*}^\top (P - P^*)\rVert_2 &= \sqrt{\sum_{i=1}^n\bigg(\sum_{j=1}^n \pi_j^*(P_{ji}-P_{ji}^*) \bigg)^2}\\
&\leq 2 \max_{v\in \mathcal{V}} \sum_{i=1}^n v_i \bigg(\sum_{j\neq i} \pi_j^*(P_{ji}-P_{ji}^*) + \pi_i^*(P_{ij}^* - P_{ij})  \bigg)\,.
\end{aligned}
\end{equation}
We now expand on the linear term:
\begin{equation*}
\begin{aligned}
&\sum_{i=1}^m v_i \bigg( \sum_{j\neq i} \pi_j^* (P_{ji} - P_{ji}^*) + \pi_i^*(P_{ij}^* - P_{ij}) \bigg)\\
&= \frac{1}{d} \sum_{i=1}^m v_i \bigg( \sum_{j\neq i} \pi_j^* \big[\sum_{l=1}^n A_{li}A_{lj}\big(X_{lj}(1-X_{li}) - \E[X_{lj}(1-X_{li})]\big) \big] \\
&\quad\quad- \sum_{j\neq i}\pi_i^* \big[\sum_{l=1}^n A_{li}A_{lj}\big(X_{li}(1-X_{lj}) - \E[X_{li}(1-X_{lj})]\big) \big] \bigg)\\
&=\frac{1}{d} \sum_{l=1}^n\sum_{i=1}^m \bigg( \sum_{j\neq i} v_i \pi_j^*A_{li}A_{lj} \big[(X_{lj}(1-X_{li}) - \E[X_{lj}(1-X_{li})]\big]\big) \\
&-\frac{1}{d}\sum_{l=1}^n\sum_{i=1}^m \bigg( \sum_{j\neq i} v_i \pi_i^*A_{li}A_{lj} \big[(X_{li}(1-X_{lj}) - \E[X_{li}(1-X_{lj})]\big]\big) \\
&=\frac{1}{d} \sum_{l=1}^n \sum_{i=1}^m \sum_{j\neq i} \bigg( (v_i - v_j)\pi_j^*A_{li}A_{lj} \big[(X_{lj}(1-X_{li}) - \E[X_{lj}(1-X_{li})]\big]\bigg)\,.
\end{aligned}
\end{equation*}

We will use the method of bounded difference to obtain a concentration inequality on the above sum. Note that this sum is essentially a function $f$ of $n\times m$ independent Bernoulli random variables $\{X_{li}\}$. Let $X$ and $X'$ be identical copies except for $X_{li}\neq X'_{li}$.
\begin{equation}\label{eqn:absolute-diff}
\begin{aligned}
\lvert f(X) - f(X')\rvert &= \frac{1}{d} \lvert \sum_{j\neq i} A_{li}A_{lj}(v_i - v_j) [X_{lj}(\pi^*_i - \pi_j^*) - \pi_i^*] \rvert \,.\\
\end{aligned}
\end{equation}
Ignore the normalization factor $d$ for now. Using Cauchy-Schwarz, we can upper bound the absolute difference term as
\begin{equation*} 
\begin{aligned}
&\sum_{j\neq i}  A_{li}A_{lj}(v_i - v_j)  \underbrace{[X_{lj}(\pi^*_i - \pi_j^*) - \pi_i^*]}_{\leq \max\{\pi^*_i, \pi^*_j\} \leq \frac{e^\kappa}{m}}\\
&\leq \frac{e^\kappa}{m} \cdot \sqrt m \cdot \sqrt{\sum_{j\neq i} A_{li}A_{lj}(v_i - v_j)^2} = \frac{e^\kappa}{\sqrt m} \cdot \sqrt{\sum_{j\neq i} A_{li}A_{lj}(v_i - v_j)^2} \,.\\
\end{aligned}
\end{equation*}
At this point we can invoke concentration inequality based on bounded difference (e.g., Hoeffding's inequality).
Fixing a $v \in \mathcal{V}$, we have
\begin{equation*}
\begin{aligned}
&\Pr\bigg(\sum_{l=1}^n \sum_{i=1}^m \sum_{j\neq i} (v_i - v_j)\pi_j^*A_{li}A_{lj} \big[(X_{lj}(1-X_{li}) - \E[X_{lj}(1-X_{li})]\big] > t\,\lvert\, \A  \bigg) \\
&\leq 2\exp{-\frac{2t^2}{\sum_{l=1}^n \sum_{i=1}^m \frac{e^{2\kappa} }{m} \cdot \sum_{j\neq i} A_{li}A_{lj}(v_i - v_j)^2   }}\\
&= 2\exp{-\frac{2t^2}{ \frac{e^{2\kappa} }{m} \sum_{i\neq j} B_{ij} (v_i - v_j)^2  }}\\
&[\text{Conditioned on $\A$, $B_{ij} \leq \frac{3np^2}{2} $}]\\
&\leq 2\exp{-\frac{2t^2}{\frac{e^{2\kappa} }{m} \sum_{i\neq j} \frac{3}{2}np^2 \cdot (v_i - v_j)^2 }}\\
&\leq 2\exp{-\frac{2t^2}{ \frac{e^{2\kappa}}{m} \frac{3}{2}mnp^2  }}\\
&= 2\exp{-\frac{4t^2}{3e^{2\kappa} np^2  }} \,.\\
\end{aligned}
\end{equation*}

Note that our $\frac{1}{2}$-net has cardinality $(\frac{2}{\frac{1}{2}} + 1)^m = 5^m$ (cf. Corollary 4.2.13 \cite{vershynin2018high}) and we are interested in the probability that large deviation \emph{doesn't happen for all $v\in \mathcal{V}$}. Applying union bound over all $v\in \mathcal{V}$, we have
\begin{equation*}
\begin{aligned}
&\Pr\bigg(\sum_{l=1}^n \sum_{i=1}^m \sum_{j\neq i} (v_i - v_j)\pi_j^*A_{li}A_{lj} \big[(X_{lj}(1-X_{li}) - \E[X_{lj}(1-X_{li})]\big] > t \quad\forall v \in \mathcal{V} \,\lvert \,\A \bigg) \\
&\leq 2\cdot 5^m \cdot \exp{ -\frac{4t^2}{3e^{2\kappa}np^2} }\\
&\leq \exp{-\frac{4t^2}{3e^{2\kappa}np^2} + 4m }\,.
\end{aligned}
\end{equation*}
Set
$$t = e^\kappa \sqrt{\frac{3np^2}{4}} \cdot \sqrt{4m + 12\max\{m, \log{np^2}\} } .$$
Then $ \exp{-\frac{4t^2}{3e^{2\kappa}np^2} + 4m } \leq \min\{\exp{-12m}, \frac{1}{(np^2)^{12}}\}$. Consequently,
$$ \lVert {\pi^*}^\top (P - P^*)\rVert_2 \leq \frac{ 2e^\kappa\sqrt{\frac{3np^2}{4}} \cdot \sqrt{16 \cdot \max\{m, \log np^2  \}}}{d} \leq \frac{2e^\kappa \sqrt{np^2}  \cdot \sqrt{12 \cdot \max\{m, \log np^2  \} }}{d} $$
with probability at least $1-  \min\{\exp{-12m}, \frac{1}{(np^2)^{12}} \} $. Note that the factor of $2$ comes from Equation (\ref{eqn:norm-linear}). Substituting $d = \frac{3mnp^2}{2}$ into the bound above finishes the proof.
\end{proof}

\subsection{Putting it all together}

The results of previous sections provide bounds on the numerator and denominator of the eigenperturbation bound in Lemma \ref{lem:perturb-bound-mc}. We can now combine all of them towards obtaining a bound on $\lVert \beta - \beta^*\rVert_2$. As an intermediate, we first obtain the following bound on $\lVert \pi - \pi^* \rVert_2$:

\begin{theorem}\label{thm:stationary-error-bound} Consider the random sampling scheme described in Section \ref{sect:error-guarantees}. Suppose that $np^2 \geq \max\{C_2 \frac{12^2 e^{4\kappa}}{\gamma^2} \log m, C_1\log m\} $ for sufficiently large constants (e.g., $C_2 \geq 30$, $C_1 \geq 101$). Then
$$ \lVert \pi - \pi^* \rVert_{2} \leq \frac{48/\sqrt 3 e^{3\kappa}}{\gamma}\cdot \frac{\sqrt{ \max\{m, \log np^2  \}  } }{m\sqrt{np^2}} $$
with probability at least $1- \min\{\exp{-12m}, \frac{1}{(np^2)^{12}} \} -  \exp{-\frac{\gamma^2np^2}{10 \cdot 12^2 \cdot e^{4\kappa}}} - \exp{-\frac{np^2}{20}} $.
\end{theorem}

\begin{proof} We first assume that $\A$ holds. The probability bound in the theorem statement can be obtained following a simple union bound argument.

From the conclusions of Lemma \ref{lem:projected-error-bound}, we have
$$  \Pr\bigg(\lVert {\pi^*}^\top (P - P^*)\rVert_2 > \frac{2e^\kappa \sqrt{16 \cdot \max\{m, \log np^2  \} }}{m \sqrt{np^2}} \,\lvert \, \A \bigg) \leq  \min\{\exp{-12m}, \frac{1}{(np^2)^{12}}\} \,. $$ 

From the conclusion of Corollary \ref{coro:spectral-error-difference}, we have
$$ \Pr\bigg(\mu^*(P) - \lVert P - P^* \rVert_2 < \frac{\gamma}{6 e^{2\kappa}} \,\lvert \, \A \bigg) \leq \exp{-\frac{\gamma^2np^2}{10 \cdot 12^2 \cdot e^{4\kappa}}}  \,. $$

Applying Lemma \ref{lem:perturb-bound-mc}: Conditioned on $\A$ and applying union bound over the two rare events above, the following holds with probability at least $1- \min\{\exp{-12m}, \frac{1}{{(np^2)}^{12}} \} -  \exp{-\frac{\gamma^2np^2}{10 \cdot 12^2 \cdot e^{4\kappa}}}$.
$$ \lVert \pi - \pi^* \rVert_{2} \leq \frac{\lVert {\pi^*}^\top(P^* - P)\rVert_{2}  }{\mu(P^*) - \lVert P - P^* \rVert_{2} } \leq \frac{48/\sqrt 3 e^{3\kappa}}{\gamma}\cdot \frac{\sqrt{ \max\{m, \log np^2  \}  } }{m\sqrt{np^2}}\,. $$
Let the above good event be $\B$.
\begin{equation*}
\begin{aligned}
\Pr(\B^c) &= \Pr(\B^c, \A) + \Pr(\B^c, \A^c)\\
&\leq \Pr(\B^c\,\lvert\,\A) \cdot \Pr(\A) + \Pr(\A^c)\\
&\leq \Pr(\B^c\,\lvert\,\A) + \Pr(\A^c)\\
&\leq  \min\{\exp{-12m}, \frac{1}{{(np^2)}^{12}} \} + \exp{-\frac{\gamma^2np^2}{10 \cdot 12^2 \cdot e^{4\kappa}}} +  \exp{-\frac{np^2}{20}}\,.
\end{aligned}
\end{equation*}
This completes the proof.
\end{proof}

With these results, we are finally ready to prove the main theorem providing error bounds on the parameters returned by our spectral algorithm.

\begin{reptheorem}{thm:parameters-error-bound} Consider the random sampling scheme described in Section \ref{sect:error-guarantees}. Suppose that $np^2 \geq \max\{C_2 \frac{12^2 e^{4\kappa}}{\gamma^2} \log m, C_1\log m\} $ for sufficiently large constants (e.g., $C_2 \geq 30$, $C_1 \geq 101$). Then the output of the spectral algorithm (Algorithm \ref{alg:spectral}) satisfies
$$ \lVert \beta - \beta^* \rVert_2 \leq \frac{96/\sqrt{3}\cdot e^{4\kappa}}{\gamma} \cdot \frac{\sqrt{ \max\{m, \log np^2  \}  } }{\sqrt{np^2}} $$
with probability at least $1- \min\{\exp{-12m}, \frac{1}{(np^2)^{12}} \} -  \exp{-\frac{\gamma^2np^2}{10 \cdot 12^2 \cdot e^{4\kappa}}} - \exp{-\frac{np^2}{20}} $.
\end{reptheorem}

\begin{proof}
Suppose for now that there is a factor $L$ such that $\lvert \log x - \log x' \rvert \leq L \lvert x - x'\rvert$ for all $x, x' \in [\pi^*_{\min}, \pi^*_{\max}]$. 
One can see that the output of the spectral algorithm is the output of the truncated log function
$$ \beta_i = \tilde{\log}(\pi_i) - \frac{1}{m} \sum_{k=1}^m \tilde\log(\pi_k) \,. $$
where $\tilde \log(\pi_i) = \log\left(\max\{ \pi_i, \frac{1}{me^{\kappa}}\}\right) $. On the other hand $\beta^*$ is related to $\pi^*$ via the same transformation.
$$ \beta^*_i = \log \pi^*_i - \frac{1}{m} \sum_{k=1}^m \log \pi^*_k = \tilde\log(\pi^*_i) - \frac{1}{m} \sum_{k=1}^m \tilde\log(\pi^*_k) \,. $$
The equality holds because by definition $\pi_i^* \geq \frac{1}{me^{\kappa}}$. The goal is to relate $\lVert \beta-\beta^*\rVert_2$ to $\lVert \pi -\pi^*\rVert_2$. 
\begin{equation*}
\begin{aligned}
\lVert \beta - \beta^* \rVert_2^2 &= \sum_{i=1}^m (\beta_i - \beta^*_i)^2 = \sum_{i=1}^m \left(\tilde\log (\pi_i) - \frac{1}{m}\sum_k \tilde\log(\pi_k) - \tilde\log(\pi_i^*) + \frac{1}{m}\sum_k \tilde\log(\pi^*_k) \right)^2 \\
&= \sum_{i=1}^m (\tilde\log(\pi_i) - \tilde\log(\pi^*_i) + \frac{1}{m}\sum_k [\tilde\log(\pi^*_k) - \tilde\log(\pi_k)] )^2\\
&\leq 2\sum_{i=1}^m \bigg((\tilde\log(\pi_i) -\tilde\log(\pi^*_i))^2 + (\frac{1}{m}\sum_k [\tilde\log(\pi^*_k) - \tilde\log(\pi_k)] )^2\bigg)\\
&= 2\sum_{i=1}^m \left((\tilde\log(\pi_i) - \tilde\log(\pi^*_i))^2 + 2m \cdot \frac{1}{m^2} \big(\sum_k [\tilde\log(\pi^*_k) - \tilde\log(\pi_k)] )^2 \right)\\
&\leq 2L^2 \sum_{i=1}^m \lvert \pi_i - \pi^*_i \rvert^2 + \frac{2}{m} \cdot m \cdot \sum_{k=1}^m (\tilde\log(\pi^*_k) - \tilde\log(\pi_k))^2\\
&\leq 2L^2 \lVert \pi - \pi^* \rVert_
2^2 + 2L^2 \sum_{k=1}^m \lvert \pi_k - \pi^*_k \rvert^2\\
&= 4L^2 \lVert \pi - \pi^* \rVert_2^2\,.
\end{aligned}
\end{equation*}
Taking the square root of both sides of the inequality gives
$$ \lVert \beta - \beta^* \rVert_2 \leq 2L \lVert \pi - \pi^* \rVert_2\,. $$
Observe that $\pi^*_{\min} \geq \frac{1}{me^{\kappa}}$. One can thus easily see that the $\tilde\log$ function within the dynamic range has gradient absolutely bounded by $me^{\kappa}$. Therefore $L \leq me^{\kappa}$. Substituting this upper bound on $L$ into the inequality obtained above and combining with the conclusion of Theorem \ref{thm:stationary-error-bound} completes the proof.
\end{proof}

\begin{repcoro}{coro:consistency} Consider the setting of Theorem \ref{thm:parameters-error-bound} and for a fixed $m$ with $p = 1$, the spectral algorithm is a consistent estimator of $\beta^*$. That is, its output $\beta$ satisfies
$\lim_{n\rightarrow \infty} \Pr(\lVert \beta - \beta^* \rVert_2 < \epsilon) = 1 \,, \forall \epsilon > 0\,.$
\end{repcoro}

\begin{proof} It is easy to see from the conclusion of Theorem \ref{thm:parameters-error-bound} that as $n\rightarrow \infty$, for a fixed $m$ and $p=1$ (or any constant $p$ for that matter), $\lVert \beta - \beta^* \rVert_2 \rightarrow 0$ and 
$$1- \min\{\exp{-12m}, \frac{1}{(n)^{12}} \} -  \exp{-\frac{\gamma^2n}{10 \cdot 12^2 \cdot e^{4\kappa}}} - \exp{-\frac{n}{20}} \rightarrow 1\,. $$
\end{proof}

We now prove the error bounds when $m$ is allowed to grow. The proof of Theorem \ref{coro:parameters-error-bound-growing-m} is almost identical to that of Theorem \ref{thm:parameters-error-bound}. The key difference is that under condition $\A^+$ (which happens with probability at least $1-n^{-9}$ given that $mp \geq C''\log n$ for a sufficiently large constant $C''$), one can obtain a stronger bound on the projected error term $\lVert {\pi^*}^\top(P - P^*)\rVert_2$. The difference in the projected term is summarized by the lemma below.

\begin{lemma}\label{lem:projected-error-bound-growing-m} Conditioned on event $\A^+$,
$$ \lVert {\pi^*}^\top (P - P^*)\rVert_2  \leq e^\kappa \sqrt{\frac{8}{mnp}} $$ 
with probability at least $1- \exp{-12m}$ over the random responses of the users.
\end{lemma}

\begin{proof} The proof here is almost identical to that of Lemma \ref{lem:projected-error-bound}. The key difference is that under conditioned $\A^+$ we could obtain better bound also using the bounded difference method. Namely, one can invoke Cauchy-Schwarz on the absolute difference term in Equation (\ref{eqn:absolute-diff}) as follows:
\begin{equation*} 
\begin{aligned}
&\sum_{j\neq i}  A_{li}A_{lj}(v_i - v_j)  \underbrace{[X_{lj}(\pi^*_i - \pi_j^*) - \pi_i^*]}_{\leq \max\{\pi^*_i, \pi^*_j\} \leq \frac{e^\kappa}{m}}\\
&\leq \frac{e^\kappa}{m}\cdot \sum_{j\neq i}  A_{li}A_{lj}(v_i - v_j) =  \frac{e^\kappa}{m}\cdot \sum_{j\neq i} A_{lj} A_{li}A_{lj}(v_i - v_j)\\
&\leq \frac{e^\kappa}{m} \cdot \sqrt{\underbrace{\sum_j A_{lj}}_{\leq \frac{3}{2}mp}} \cdot \sqrt{\sum_{j\neq i} A_{li}A_{lj}(v_i - v_j)^2} = \sqrt{3/2} \cdot \frac{e^\kappa\sqrt p}{\sqrt m} \cdot \sqrt{\sum_{j\neq i} A_{li}A_{lj}(v_i - v_j)^2}\,.\\
\end{aligned}
\end{equation*}
Continuting the same procedures as in the proof of Lemma \ref{lem:projected-error-bound} gives
\begin{equation*}
\begin{aligned}
&\Pr\bigg(\sum_{l=1}^n \sum_{i=1}^m \sum_{j\neq i} (v_i - v_j)\pi_j^*A_{li}A_{lj} \big[(X_{lj}(1-X_{li}) - \E[X_{lj}(1-X_{li})]\big] > t \quad\forall v \in \mathcal{V} \,\lvert \,\A \bigg) \\
&\leq 2\cdot 5^m \cdot \exp{-\frac{2t^2}{\sum_{l=1}^n \sum_{i=1}^m \frac{e^{2\kappa} }{m} \cdot \sum_{j\neq i} A_{li}A_{lj}(v_i - v_j)^2   }}\\
&= 2\cdot 5^m\cdot\exp{-\frac{2t^2}{ \frac{3e^{2\kappa} p }{2m} \sum_{i\neq j} B_{ij} (v_i - v_j)^2  }}\\
&\leq 2\cdot 5^m \cdot \exp{ -\frac{8t^2}{9e^{2\kappa}p \cdot np^3} }\\
&\leq \exp{-\frac{8t^2}{9e^{2\kappa}np^3} + 4m }\,.
\end{aligned}
\end{equation*}
Set
$$t = e^\kappa \sqrt{\frac{9np^3}{8}} \cdot \sqrt{16m} .$$
Then $ \exp{-\frac{8t^2}{9e^{2\kappa}np^3} + 4m } \leq \exp{-12m}$. Consequently,
$$ \lVert {\pi^*}^\top (P - P^*)\rVert_2 \leq \frac{ e^\kappa\sqrt{ 18 mnp^3 }   }{d}  $$
with probability at least $1-  \exp{-12m} $.  Substituting $d = \frac{3mnp^2}{2}$ into the bound above finishes the proof.
\end{proof}

With the numerator the eigenperturbation bound in Lemma \ref{lem:perturb-bound-mc} updated, we now have the proof for Theorem \ref{coro:parameters-error-bound-growing-m}.

\begin{reptheorem}{coro:parameters-error-bound-growing-m} Consider the random sampling scheme described in Section \ref{sect:error-guarantees}. Suppose that $np^2 \geq \max\{C_2 \frac{12^2 e^{4\kappa}}{\gamma^2} \log m, C_1\log m\} $ and $mp \geq C''\log m$ for sufficiently large constants $C_1, C_2, C''$ (e.g., $C_2 \geq 30$, $C_1, C'' \geq 101 $). Then the output of the spectral algorithm (Algorithm \ref{alg:spectral}) satisfies
$$ \lVert \beta - \beta^* \rVert_2 \leq \frac{96/\sqrt{2} e^{4\kappa}}{\gamma} \cdot \frac{\sqrt m}{\sqrt{np}}  $$
with probability at least $1- \exp{-12m} - \exp{-\frac{\gamma^2np^2}{10 \cdot 12^2 \cdot e^{4\kappa}}} - n^{-9}$.
\end{reptheorem}

\begin{proof} Applying Lemma \ref{lem:perturb-bound-mc} and Corollary \ref{coro:spectral-error-difference} gives
$$ \lVert \pi - \pi^* \rVert_2 \leq \frac{48/\sqrt{2} e^{3\kappa}}{\gamma} \cdot \frac{1}{\sqrt{mnp}} \,.$$
Following the same proof as that of Theorem \ref{thm:parameters-error-bound} with minor changes to the constant factor completes the proof.
\end{proof}

\newpage
\section{Proofs of Lower Bounds}

\begin{reptheorem}{thm:cramer-rao-lower-bound-known-theta} Consider the sampling model described in in Section \ref{sect:error-guarantees}. Let $T$ be any unbiased estimator for the item parameters. Then the mean squared error of such estimator is lower bounded as
$$ \E\lVert \hat\beta - \beta^*\lVert^2_2 \, \geq \frac{4m}{np} \,. $$
\end{reptheorem}

\begin{proof} Suppose that we know exactly the user parameter $\theta^*$. It is known that the presence of unknown nuisance parameters does not make the estimation problem easier (cf. \cite[pp. 127-128]{lehmann2006theory}). On the other hand, once the user parameters are exactly known, the $\beta^*$ estimation problems reduces to $m$ estimation problems over each parameter $\beta^*_i$ for $i \in [m]$. The Fisher information for a single parameter $I(\beta^*_i)$ is
\begin{equation*}
\begin{aligned}
I(\beta^*_i) &= -\sum_{l=1}^n \E_{X_{li}}\bigg[ \bigg(\frac{\partial \ell(X_{li}; \beta^*_i)}{\partial \beta^*_i}\bigg)^2 \bigg]\\
&= -\sum_{l=1}^n \ p \bigg[\frac{1}{1+e^{-(\theta^*_l - \beta^*_i)}} \frac{\partial^2}{\partial\beta_i^2} \log\bigg(\frac{1}{1+e^{-(\theta^*_l - \beta^*_i)}}\bigg)  + \frac{1}{1+e^{-(\beta^*_i - \theta^*_l)}} \frac{\partial^2}{\partial\beta_i^2} \log\bigg(\frac{1}{1+e^{-(\beta^*_i - \theta^*_l)}}\bigg)   \bigg]  \\
&= \sum_{l=1}^n  p \bigg[\frac{e^{-(\theta^*_l - \beta^*_i)}}{(1+e^{-(\theta^*_l - \beta^*_i)})^2 }  \bigg]\\
&\leq \frac{np}{4} \,.
\end{aligned}
\end{equation*}
The last inequality comes from the observation that $\frac{e^{-(\theta - \beta^*_i)}}{(1+e^{-(\theta - \beta^*_i)})^2 } \leq \frac{1}{4} \,\forall \theta \in \R$.
% where $\omega' = \min_{l, i} \{ \frac{1}{1+e^{-(\theta^*_l - \beta^*_i)}} \cdot \frac{1}{1+e^{-(\beta^*_i - \theta^*_l)}} \}$. Intuitively, this is the minimum variance for a user response $X_{li}$. Thanks to the concave nature of the function $x(1-x)$ for $x\in (0,1)$, we can lower bound $\omega'$ under parameter bounded assumption as follows:
% \begin{equation*}
% \begin{aligned}
% \omega' \geq \min\bigg\{ &\frac{1}{1+e^{-(\theta^*_{\min} - \beta^*_{\max} )}} \cdot \frac{1}{1+e^{-(\beta^*_{\max} - \theta^*_{\min})}},\\ 
%     & \frac{1}{1+e^{-(\theta^*_{\max} - \beta^*_{\min} )}} \cdot \frac{1}{1+e^{-(\beta^*_{\min} - \theta^*_{\max})}}   \bigg\}
% \end{aligned}
% \end{equation*}
% The $\omega$ constant in the proof is simply $1/\omega'$. This finishes the proof.
Repeating the same argument for every item parameter $i\in [m]$ and applying the Cramer-Rao lower bound for multivariate parameter finishes the proof.
\end{proof}

\begin{reptheorem}{thm:cramer-rao-m2} Fix $m=2$ and consider the sampling model described in in Section \ref{sect:error-guarantees}. Let $T$ be any unbiased estimator for the user parameters that only uses \emph{pairwise differential measurements}. Then the mean squared error of such estimator is lower bounded as
$$ \E\lVert T(X) - \beta^*\lVert^2_2 \,\geq \frac{1}{np^2} \,, $$
where $T(X)$ is the output of the estimator $T$ when given data $X$.
\end{reptheorem}

\begin{proof} Since there are only two parameters $\beta^*_1$ and $\beta^*_2$ and $\beta_1^* + \beta_2^* = 0$, the problem reduces to estimating a single parameter $\delta^*$ where $\beta^*_1 = \delta^*$, $\beta^*_2 = -\delta^*$. To establish the result in the theorem statement, we first need to introduce a modified likelihood function that adapts to the setting where we consider the class of algorithms that uses pairwise differential measurements. 

Note that for $m=2$, an observation $x \in \{(1,0), (0,1), (1,*), (0, *), (*,1), (*,0), (*,*) \}$. To prove the lower bound, we need to define an alternative observation model (and with it an alternative likelihood) appopriate to the class of estimators that only use pairwise differential measurements. Define the pseudo-observation $x'$ (in terms of $x$) as
$$ x' = \begin{cases} (1,0) &\text{if } x = (1,0)\\ (0,1) &\text{if } x= (0,1) \\ (*,*) &\text{otherwise} \end{cases} \quad ,$$
We can show the Fisher information under the modified likelihood.
\begin{equation*}
\begin{aligned}
I(\delta^*) &= - \E\bigg[\frac{\partial^2}{\partial\delta^2} \log \tilde L(X', \delta^*)  \bigg]\\
&= - \sum_{l=1}^n \bigg[ \Pr(x_l' = (1,0)) \cdot \frac{\partial^2}{\partial\delta^2} \log \Pr(x_l'=(1,0)) +  \Pr(x_l' = (0,1)) \cdot \frac{\partial^2}{\partial\delta^2} \log \Pr(x_l'=(0,1)) \bigg]\\
&= -p^2 \sum_{l=1}^n \bigg[ \frac{1}{1+e^{-(\theta^*_l -\delta^*)}} \cdot \frac{1}{1+e^{-(-\delta^* -\theta^*_l)}} \cdot \frac{\partial^2}{\partial\delta^2} \log\bigg(\frac{1}{1+e^{-(\theta^*_l -\delta^*)}} \cdot \frac{1}{1+e^{\delta^* + \theta^*_l}}\bigg) \\
&  +   \frac{1}{1+e^{-(\delta^* - \theta^*_l)}} \cdot \frac{1}{1+e^{-(\theta^*_l -(-\delta^*) ) }} \cdot \frac{\partial^2}{\partial\delta^2} \log\bigg(\frac{1}{1+e^{\theta^*_l -\delta^*}} \cdot \frac{1}{1+e^{-\delta^* - \theta^*_l}}\bigg)\bigg ]\\
&= -p^2 \sum_{l=1}^n \bigg[ \frac{1}{1+e^{-(\theta^*_l -\delta^*)}} \cdot \frac{1}{1+e^{-(-\delta^* -\theta^*_l)}} \cdot (-\frac{e^{\theta^*_l+\delta^*}}{(1+e^{\theta^*_l+\delta^*})^2} - \frac{e^{-(\theta^*_l-\delta^*)}}{(1+ e^{-(\theta^*_l-\delta^*)} )^2}) \\
&+  \frac{1}{1+e^{-(\delta^* - \theta^*_l)}} \cdot \frac{1}{1+e^{-(\theta^*_l +\delta^*) }} \cdot (-\frac{e^{-(\theta^*_l+\delta^*)}}{(1+e^{-(\theta^*_l+\delta^*)})^2} - \frac{e^{(\theta^*_l-\delta^*)}}{(1+ e^{(\theta^*_l-\delta^*)} )^2})\bigg]\\
&= p^2 \sum_{l=1}^n \bigg[ \frac{1}{1+e^{-(\theta^*_l -\delta^*)}} \cdot \frac{1}{1+e^{\delta^* +\theta^*_l}} \cdot (\frac{e^{\theta^*_l+\delta^*}}{(1+e^{\theta^*_l+\delta^*})^2} +\frac{e^{-(\theta^*_l-\delta^*)}}{(1+ e^{-(\theta^*_l-\delta^*)} )^2}) \\
&+  \frac{1}{1+e^{-(\delta^* - \theta^*_l)}} \cdot \frac{1}{1+e^{-(\theta^*_l +\delta^*) }} \cdot (\frac{e^{-(\theta^*_l+\delta^*)}}{(1+e^{-(\theta^*_l+\delta^*)})^2} + \frac{e^{(\theta^*_l-\delta^*)}}{(1+ e^{(\theta^*_l-\delta^*)} )^2})\bigg]\\
&\leq np^2 \,.
\end{aligned}
\end{equation*}
This finishes the proof.
\end{proof}

\newpage
\section{Proofs of Results in Section \ref{sect:accelerated}}

\begin{reptheorem}{thm:equivalent-markov-chains}
Consider the modified Markov chain $\bar P$ constructed per Equation (\ref{eqn:accelerated-mc}) and the original Markov chain $P$ constructed per Equation (\ref{eqn:emp-markov-chain-def}). Suppose that $\bar P$ and $P$ admit unique stationary distributions $\bar\pi$ and $\pi$, respectively. Then
$$ \bar\pi_i = \frac{\pi_i d_i}{\sum_{k=1}^m\pi_k d_k  } \quad \forall i\,,$$
where $d_i$ are the normalization factors in the construction of the modified Markov chain $\bar P$.
\end{reptheorem}

\begin{proof} 
As $\bar \pi$ is the unique stationary distribution of the modified Markov chain $\bar P$, it must satisfy the fixed point equation:
$$ \sum_{k\neq i} \bar\pi_k \bar P_{ki} = \sum_{k\neq i} \bar \pi_i \bar P_{ik} \quad\forall i \in [m]\,.$$
Intuitively, this means the total 'inflow' into a state $i$ is equal to the total 'outflow' out of state $i$. By construction, this is equivalent to
$$ \sum_{k\neq i} \bar\pi_k \frac{Y_{ik}}{d_k}  = \sum_{k\neq i} \bar \pi_i \frac{Y_{ik}}{d_i} \quad\forall i \in [m] \quad (*)\,.$$
Similarly, we have for the original Markov chain $P$,
$$ \sum_{k\neq i} \pi_k Y_{ki} = \sum_{k\neq i} \pi_i Y_{ik} \quad\forall i \in [m]\,.$$
Note that in the original Markov chain, because we use a global normalization constant $d$, it does not appear in the fixed point equation. 
One can easily see that setting $\bar \pi_i = \frac{\pi_i d_i}{\sum_{k=1}^m \pi_k d_k}$ preserves the fixed point equation in $(*)$. Assuming that both Markov Chains admit unique stationary distributions, there is thus a 1-1 relation between $\bar \pi$ and $\pi$ as stated in the theorem. This finishes the proof.
\end{proof}

\newpage
\section{Additional Experiments}

\subsection{Other Pairwise Methods in the Literature}

In this section we describe three methods that are related to our algorithm. As noted before, previous matrix methods in the literature construct an item-item matrix and assumes that such matrix is dense. It is unclear how one would generalize these methods to the case where the item-item matrix is sparse, which is quite commonly observed in real life datasets. A common quantity that the previous matrix methods use is 
$$ f_{ij} = \Pr_{l\in [n]} (X_{li}=1, X_{lj}=0 \,\lvert \, X_{li} + X_{lj}=1 )\,.$$
Intuitively, $f_{ij}$ is the empirical probability at which a user responds $1$ to item $i$ and $0$ to item $j$, conditioned on the event that the user responds to exactly only one of the two items. Suppose that we have collected $f_{ij}$ for all $i\neq j$, consider a matrix $D$ defined entrywise as
$$ D_{ij} = \frac{f_{ji}}{f_{ij}}\,. $$
This is also known as the \emph{positive reciprocal matrix}.

\textbf{The Row Sum Approach of Choppin \cite{choppin1982fully}:} Given the matrix $D$, construct a matrix $\ln D$ by taking the $\log$ of every entry of $D$. The row sums of this $\ln D$ matrix is (after appropriate normalization) produce an estimate $\beta$. To see why, it is helpful to first check that in the limit of infinite data and suppose that we observe all pairs $i, j$ then $f_{ij}$ is exact and
$$ f_{ij} = \frac{e^{\beta_j^*}}{e^{\beta_i^*} + e^{\beta_j^*} }\,. $$
Then
$$ D_{ij} = \frac{e^{\beta_i^*}}{e^{\beta_j^*}}$$
and
$$ \ln D_{ij} = \beta_i^* - \beta^*_j\,.$$
It is easy to see that the row sums correspond exactly to $\beta^*$.

\textbf{The Eigenvector Method of Garner \cite{garner2002eigenvector} and Saaty \cite{saaty1987analytic}:} Given the matrix $D$, right the leading eigen-vector. Modulo an appropriate scaling factor, the leading left eigen-vector is an estimate for $e^{\beta}$. Taking the log of this eigenvector recovers $\beta$. To see why, verify that
$$ D \begin{bmatrix} e^{\beta^*_1}\\\vdots \\ e^{\beta_m^*} \end{bmatrix} = m \cdot \begin{bmatrix} e^{\beta^*_1}\\\vdots \\ e^{\beta_m^*} \end{bmatrix}\,. $$

\textbf{Pairwise Maximum Likelihood Estimate (PMLE):} Another pairwise approach used in the literature is the Pairwise Maximumum Likelihood Estimate (PMLE) \cite{zwinderman1995pairwise}. Similar to the intuition behind the pairwise methods mentioned above, PMLE uses the fact that the conditional probability $ \Pr_{l\in [n]} (X_{li}=1, X_{lj}=0 \,\lvert \, X_{li} + X_{lj}=1 )$ does not involve the user parameter. PMLE maximizes the pairwise conditional likelihood. We are not able to find any open source python implementation of PMLE so we use the majorization-minorization (MM) algorithm for estimation \cite{hunter2004tutorial} and adapt our implementation from an open source implementation \cite{choix}. We also implement a different version of PMLE using Scipy's optimization subroutine \cite{2020SciPy-NMeth}. However, the later version has very significant numerical issues and gives inaccurate results. We therefore use the MM-based version in our experiments.

\begin{figure}[h]
  \centering
  \includegraphics[height=60mm, width=80mm]{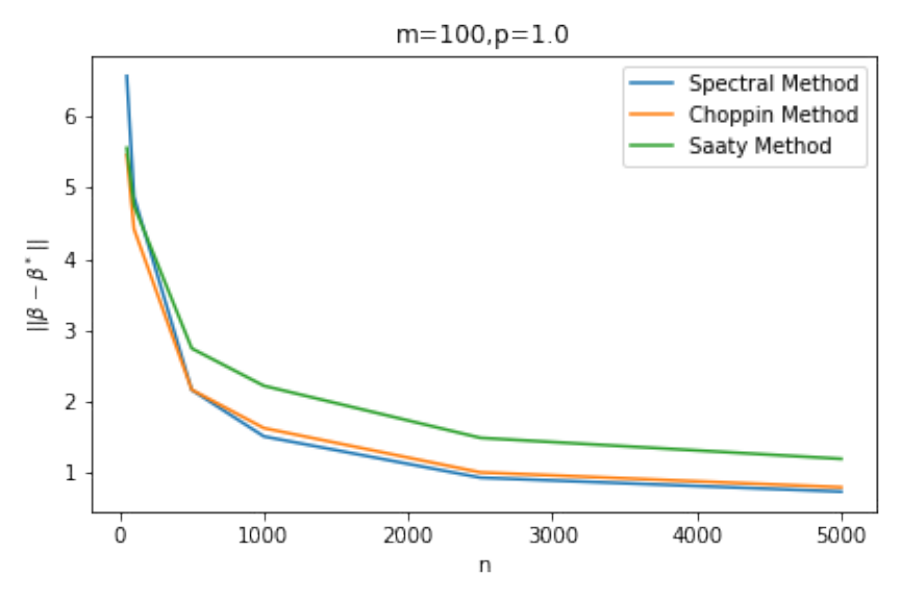}
  \caption{Comparison between our spectral method and two matrix methods in the literature. The performance are quite similar. However, our methods can be easily generalizable to the setting where the pairwise comparison matrix contains missing entries whereas the spectral methods in the literature assumes a full comparison matrix. \label{fig:eigen-methods-comparisons}}
\end{figure}

For completeness, we conduct extra experiments comparing between our spectral method and the two previously studied matrix methods on synthetic data with $m=100$ under full observation data. The result is presented in Figure \ref{fig:eigen-methods-comparisons}.

\textbf{Extra Experiment Results with Pairwise MLE:} For completeness, we also conduct extra experiments comparing between the spectral method and PMLE. The result (together with previous results reported in the main paper) is summarized in Table \ref{tbl:all-results-pair}.

\begin{table}[h!]
\begin{adjustwidth}{-2.35cm}{}
\resizebox{1.25\textwidth}{!}{%
\begin{tabular}{|c|ccccc|ccccc|ccccc|ccccc|}
\hline
          & \multicolumn{5}{c|}{AUC}  & \multicolumn{5}{c|}{Log likelihood}   & \multicolumn{5}{c|}{Top-K accuracy}  & \multicolumn{5}{c|}{Inference time}  \\ \hline
Dataset   & \multicolumn{1}{c|}{Spectral} & \multicolumn{1}{c|}{MMLE} & \multicolumn{1}{c|}{CMLE} & \multicolumn{1}{c|}{PMLE}  & JMLE & \multicolumn{1}{c|}{Spectral} & \multicolumn{1}{c|}{MMLE} & \multicolumn{1}{c|}{CMLE} & \multicolumn{1}{c|}{PMLE} & JMLE & \multicolumn{1}{c|}{Spectral} & \multicolumn{1}{c|}{MMLE} & \multicolumn{1}{c|}{CMLE} &\multicolumn{1}{c|}{PMLE} & JMLE & \multicolumn{1}{c|}{Spectral} & \multicolumn{1}{c|}{MMLE} & \multicolumn{1}{c|}{CMLE} & \multicolumn{1}{c|}{PMLE} & JMLE \\ \hline
 {LSAT} & \multicolumn{1}{c|}{$0.707$} & \multicolumn{1}{c|}{$0.707$} & \multicolumn{1}{c|}{$0.707$} & \multicolumn{1}{c|}{$0.707$}  & {$0.707$} & \multicolumn{1}{c|}{$-0.487$} & \multicolumn{1}{c|}{$-0.489$} & \multicolumn{1}{c|}{$-0.487$} & \multicolumn{1}{c|}{$-0.487$} & {$-0.485$}  & \multicolumn{1}{c|}{N/A} & \multicolumn{1}{c|}{N/A} & \multicolumn{1}{c|}{N/A} & \multicolumn{1}{c|}{N/A} & {N/A} & \multicolumn{1}{c|}{$0.028$} & \multicolumn{1}{c|}{$0.159$} & \multicolumn{1}{c|}{$0.154$} & \multicolumn{1}{c|}{$0.011$} & {$0.075$} \\ \hline
 {UCI} & \multicolumn{1}{c|}{$0.565$} & \multicolumn{1}{c|}{$0.565$} & \multicolumn{1}{c|}{$0.565$} & \multicolumn{1}{c|}{$0.565$}  & {$0.565$} & \multicolumn{1}{c|}{$-0.687$} & \multicolumn{1}{c|}{$-0.686$} & \multicolumn{1}{c|}{$-0.692$} & \multicolumn{1}{c|}{$-0.687$} & {$-0.706$}  & \multicolumn{1}{c|}{N/A} & \multicolumn{1}{c|}{N/A} & \multicolumn{1}{c|}{N/A} & \multicolumn{1}{c|}{N/A} & {N/A} & \multicolumn{1}{c|}{$0.015$} & \multicolumn{1}{c|}{$0.133$} & \multicolumn{1}{c|}{$0.136$} & \multicolumn{1}{c|}{$0.015$} & {$0.034$} \\ \hline
 {3 GRADES} & \multicolumn{1}{c|}{$0.532$} & \multicolumn{1}{c|}{$0.532$} & \multicolumn{1}{c|}{$0.532$} & \multicolumn{1}{c|}{$0.532$}  & {$0.532$} & \multicolumn{1}{c|}{$-0.706$} & \multicolumn{1}{c|}{$-0.692$} & \multicolumn{1}{c|}{$-0.699$} & \multicolumn{1}{c|}{$-0.704$} & {$-0.717$}  & \multicolumn{1}{c|}{N/A} & \multicolumn{1}{c|}{N/A} & \multicolumn{1}{c|}{N/A} & \multicolumn{1}{c|}{N/A} & {N/A} & \multicolumn{1}{c|}{$0.021$} & \multicolumn{1}{c|}{$0.181$} & \multicolumn{1}{c|}{$0.105$} & \multicolumn{1}{c|}{$0.011$} & {$0.009$} \\ \hline
 {RIIID} & \multicolumn{1}{c|}{$0.723$} & \multicolumn{1}{c|}{$0.724$} & \multicolumn{1}{c|}{N/A} & \multicolumn{1}{c|}{$0.724$}  & {$0.724$} & \multicolumn{1}{c|}{$-0.486$} & \multicolumn{1}{c|}{$-0.49$} & \multicolumn{1}{c|}{N/A} & \multicolumn{1}{c|}{$-0.486$} & {$-0.486$}  & \multicolumn{1}{c|}{N/A} & \multicolumn{1}{c|}{N/A} & \multicolumn{1}{c|}{N/A} & \multicolumn{1}{c|}{N/A} & {N/A} & \multicolumn{1}{c|}{$13.1$} & \multicolumn{1}{c|}{$104$} & \multicolumn{1}{c|}{N/A} & \multicolumn{1}{c|}{$16.3\text{K}$} & {$61.2$} \\ \hline
 {HETREC} & \multicolumn{1}{c|}{$0.728$} & \multicolumn{1}{c|}{$0.729$} & \multicolumn{1}{c|}{$0.506$} & \multicolumn{1}{c|}{$0.727$}  & {$0.73$} & \multicolumn{1}{c|}{$-0.604$} & \multicolumn{1}{c|}{$-0.603$} & \multicolumn{1}{c|}{$-1.119$} & \multicolumn{1}{c|}{$-0.603$} & {$-0.602$}  & \multicolumn{1}{c|}{$0.5 \git 0.64 \git 0.6$} & \multicolumn{1}{c|}{$0.0 \git 0.0 \git 0.02$} & \multicolumn{1}{c|}{$0.0 \git 0.0 \git 0.0$} & \multicolumn{1}{c|}{$0.5 \git 0.64 \git 0.58$} & {$0.0 \git 0.0 \git 0.02$} & \multicolumn{1}{c|}{$50.1$} & \multicolumn{1}{c|}{$140$} & \multicolumn{1}{c|}{$224\text{K}$} & \multicolumn{1}{c|}{$4.25\text{K}$} & {$144$} \\ \hline
 {ML-100K} & \multicolumn{1}{c|}{$0.662$} & \multicolumn{1}{c|}{$0.659$} & \multicolumn{1}{c|}{$0.498$} & \multicolumn{1}{c|}{$0.662$}  & {$0.665$} & \multicolumn{1}{c|}{$-0.646$} & \multicolumn{1}{c|}{$-0.66$} & \multicolumn{1}{c|}{$-1.159$} & \multicolumn{1}{c|}{$-0.645$} & {$-0.653$}  & \multicolumn{1}{c|}{$0.4 \git 0.6 \git 0.54$} & \multicolumn{1}{c|}{$0.0 \git 0.0 \git 0.0$} & \multicolumn{1}{c|}{$0.0 \git 0.0 \git 0.0$} & \multicolumn{1}{c|}{$0.4 \git 0.6 \git 0.5$} & {$0.0 \git 0.0 \git 0.0$} & \multicolumn{1}{c|}{$1.39$} & \multicolumn{1}{c|}{$16.2$} & \multicolumn{1}{c|}{$9.56\text{K}$} & \multicolumn{1}{c|}{$368$} & {$21$} \\ \hline
 {ML-1M} & \multicolumn{1}{c|}{$0.698$} & \multicolumn{1}{c|}{$0.701$} & \multicolumn{1}{c|}{$0.468$} & \multicolumn{1}{c|}{$0.699$}  & {$0.7$} & \multicolumn{1}{c|}{$-0.626$} & \multicolumn{1}{c|}{$-0.632$} & \multicolumn{1}{c|}{$-1.166$} & \multicolumn{1}{c|}{$-0.627$} & {$-0.63$}  & \multicolumn{1}{c|}{$0.8 \git 0.72 \git 0.72$} & \multicolumn{1}{c|}{$0.6 \git 0.6 \git 0.62$} & \multicolumn{1}{c|}{$0.0 \git 0.0 \git 0.0$} & \multicolumn{1}{c|}{$0.4 \git 0.6 \git 0.6$} & {$0.5 \git 0.64 \git 0.66$} & \multicolumn{1}{c|}{$19.2$} & \multicolumn{1}{c|}{$86.9$} & \multicolumn{1}{c|}{$156\text{K}$} & \multicolumn{1}{c|}{$1.38\text{K}$} & {$194$} \\ \hline
 {EACH MOVIE} & \multicolumn{1}{c|}{$0.716$} & \multicolumn{1}{c|}{$0.718$} & \multicolumn{1}{c|}{$0.522$} & \multicolumn{1}{c|}{$0.715$}  & {$0.716$} & \multicolumn{1}{c|}{$-0.615$} & \multicolumn{1}{c|}{$-0.613$} & \multicolumn{1}{c|}{$-0.946$} & \multicolumn{1}{c|}{$-0.616$} & {$-0.614$}  & \multicolumn{1}{c|}{$0.8 \git 0.76 \git 0.82$} & \multicolumn{1}{c|}{$0.8 \git 0.68 \git 0.84$} & \multicolumn{1}{c|}{$0.0 \git 0.0 \git 0.02$} & \multicolumn{1}{c|}{$0.7 \git 0.72 \git 0.78$} & {$0.6 \git 0.6 \git 0.72$} & \multicolumn{1}{c|}{$11.3$} & \multicolumn{1}{c|}{$329$} & \multicolumn{1}{c|}{$220\text{K}$} & \multicolumn{1}{c|}{$446$} & {$1.9\text{K}$} \\ \hline
 {ML-10M} & \multicolumn{1}{c|}{$0.714$} & \multicolumn{1}{c|}{$0.716$} & \multicolumn{1}{c|}{N/A} & \multicolumn{1}{c|}{$0.714$}  & {$0.716$} & \multicolumn{1}{c|}{$-0.617$} & \multicolumn{1}{c|}{$-0.619$} & \multicolumn{1}{c|}{N/A} & \multicolumn{1}{c|}{$-0.62$} & {$-0.618$}  & \multicolumn{1}{c|}{$0.5 \git 0.84 \git 0.7$} & \multicolumn{1}{c|}{$0.1 \git 0.28 \git 0.32$} & \multicolumn{1}{c|}{N/A} & \multicolumn{1}{c|}{$0.5 \git 0.72 \git 0.72$} & {$0.0 \git 0.32 \git 0.36$} & \multicolumn{1}{c|}{$821$} & \multicolumn{1}{c|}{$3.93\text{K}$} & \multicolumn{1}{c|}{N/A} & \multicolumn{1}{c|}{$9.53\text{K}$} & {$6.55\text{K}$} \\ \hline
 {ML-20M} & \multicolumn{1}{c|}{$0.709$} & \multicolumn{1}{c|}{$0.71$} & \multicolumn{1}{c|}{N/A} & \multicolumn{1}{c|}{$0.709$}  & {$0.71$} & \multicolumn{1}{c|}{$-0.619$} & \multicolumn{1}{c|}{$-0.619$} & \multicolumn{1}{c|}{N/A} & \multicolumn{1}{c|}{$-0.621$} & {$-0.619$}  & \multicolumn{1}{c|}{$0.5 \git 0.8 \git 0.64$} & \multicolumn{1}{c|}{$0.3 \git 0.44 \git 0.4$} & \multicolumn{1}{c|}{N/A} & \multicolumn{1}{c|}{$0.4 \git 0.6 \git 0.5$} & {$0.1 \git 0.4 \git 0.4$} & \multicolumn{1}{c|}{$1.58\text{K}$} & \multicolumn{1}{c|}{$5.36\text{K}$} & \multicolumn{1}{c|}{N/A} & \multicolumn{1}{c|}{$12.8\text{K}$} & {$4.42\text{K}$} \\ \hline
 {BX} & \multicolumn{1}{c|}{$0.546$} & \multicolumn{1}{c|}{$0.577$} & \multicolumn{1}{c|}{$0.503$} & \multicolumn{1}{c|}{$0.546$}  & {$0.57$} & \multicolumn{1}{c|}{$-0.618$} & \multicolumn{1}{c|}{$-0.612$} & \multicolumn{1}{c|}{$-0.8$} & \multicolumn{1}{c|}{$-0.627$} & {$-0.617$}  & \multicolumn{1}{c|}{$0.3 \git 0.16 \git 0.16$} & \multicolumn{1}{c|}{$0.3 \git 0.24 \git 0.2$} & \multicolumn{1}{c|}{$0.0 \git 0.0 \git 0.02$} & \multicolumn{1}{c|}{$0.3 \git 0.28 \git 0.3$} & {$0.3 \git 0.2 \git 0.18$} & \multicolumn{1}{c|}{$205$} & \multicolumn{1}{c|}{$2.02\text{K}$} & \multicolumn{1}{c|}{$156\text{K}$} & \multicolumn{1}{c|}{$338$} & {$481$} \\ \hline
 {BOOK-GENOME} & \multicolumn{1}{c|}{$0.658$} & \multicolumn{1}{c|}{$0.665$} & \multicolumn{1}{c|}{N/A} & \multicolumn{1}{c|}{$0.657$}  & {$0.654$} & \multicolumn{1}{c|}{$-0.651$} & \multicolumn{1}{c|}{$-0.645$} & \multicolumn{1}{c|}{N/A} & \multicolumn{1}{c|}{$-0.649$} & {$-0.651$}  & \multicolumn{1}{c|}{$0.6 \git 0.44 \git 0.42$} & \multicolumn{1}{c|}{$0.3 \git 0.32 \git 0.34$} & \multicolumn{1}{c|}{N/A} & \multicolumn{1}{c|}{$0.3 \git 0.44 \git 0.36$} & {$0.2 \git 0.24 \git 0.38$} & \multicolumn{1}{c|}{$2.53\text{K}$} & \multicolumn{1}{c|}{$2.56\text{K}$} & \multicolumn{1}{c|}{N/A} & \multicolumn{1}{c|}{$7.8\text{K}$} & {$4.34\text{K}$} \\ \hline
\end{tabular}}
\end{adjustwidth}
\caption{Results from Table \ref{tbl:all-results-main} with PMLE. PMLE is quite competitive when applied small datasets. However, similarly to CMLE, it tends to converge quite slowly when applied to large datasets. Overall, both PMLE and the spectral method are quite competitive but the spectral method is significantly faster. \label{tbl:all-results-pair}}
\end{table}

\textbf{Extra experiments with a Bayesian method.} All of the estimation algorithms considered so far are point estimation algorithm (i.e., return a single parameter estimate). In certain applications, one may prefer a Bayesian estimation algorithm that returns a \emph{distribution} over the estimate. Recently, Bayesian algorithms based on variational inference has received considerable attention in the IRT literature. We conduct extra experiments using the algorithm proposed in \cite{natesan2016bayesian} of which implementation can be found in \cite{lalor2019emnlp,rodriguez2021evaluation}. Table \ref{tbl:experiments-bayesian} summarizes the results on a restricted subset of experiments. One can see that the Bayesian algorithm is somewhat more accurate than the spectral algorithm. However, it is considerably more complicated and as a result runs much slower than the spectral algorithm. 

\begin{table}[]
\centering
% \begin{adjustwidth}{}{}
\resizebox{0.5\textwidth}{!}{%
\begin{tabular}{|c|c|c|}
\hline
\textbf{Dataset} & \textbf{AUC (Bayesian)}    & \textbf{AUC (Spectral)}    \\ \hline
LSAT             & 0.706                      & 0.707                      \\ \hline
3 Grades         & 0.5322                     & 0.532                      \\ \hline
UCI              & 0.565                      & 0.565                      \\ \hline
ML-100K          & 0.695                      & 0.662                      \\ \hline
                 & \textbf{LogLik (Bayesian)} & \textbf{Loglik (Spectral)} \\ \hline
LSAT             & -0.487                     & -0.487                     \\ \hline
3 Grades         & -0.681                     & -0.687                     \\ \hline
UCI              & -0.693                     & -0.706                     \\ \hline
ML-100K          & -0.646                     & -0.646                     \\ \hline
                 & \textbf{Top-K (Bayesian)}  & \textbf{Top-K (Spectral)}  \\ \hline
ML-100K          & 0; 0; 0.04;                & 0.4; 0.6; 0.54             \\ \hline
                 & \textbf{Time (Bayesian)}   & \textbf{Time (Spectral)}   \\ \hline
LSAT             & 63                         & 0.028                      \\ \hline
3 Grades         & 27                         & 0.015                      \\ \hline
UCI              & 26                         & 0.021                      \\ \hline
ML-100K          & 6700                       & 2                          \\ \hline
\end{tabular}}
\caption{While the Bayesian algorithm is somewhat more accurate than the spectral algorithm, it is considerably slower. In fact, it is significantly slower than CMLE, the slowest method considered in our main experiments. \label{tbl:experiments-bayesian}}
\end{table}

\subsection{Datasets Metadata and Experiment Setup}
Table \ref{tbl:datasets-metadata} summarizes the metadata for all the real-life datasets used in our experiments.

\begin{table}[h!]
\centering
\begin{tabular}{|c|c|c|c|}
\hline
Dataset     & $m$ & $m$ & Reference \\ \hline
LSAT        & 5   & 1000 & \cite{mcdonald2013test} \\ \hline
UCI         & 4   & 131 &  \cite{hussain2018educational} \\ \hline
3 GRADES    & 3   & 648 & \cite{cortez2008using} \\ \hline
RIIID       & 6311 & 22906 & \cite{riiid} \\ \hline
HETREC      & 10197 & 2113 & \cite{Cantador:RecSys2011} \\ \hline
ML-100K     & 1682 & 943 & \cite{harper2015movielens} \\ \hline
ML-1M       & 3952 & 6040 & \cite{harper2015movielens} \\ \hline
EACH MOVIE  & 1628 & 72916 & \cite{harper2015movielens}          \\ \hline
ML-10M      & 10681 & 71567 & \cite{harper2015movielens}          \\ \hline
ML-20M      & 27278 & 138493 & \cite{harper2015movielens} \\ \hline
BX          & 6185 & 278858 & \cite{ziegler2005improving}          \\ \hline
BOOK-GENOME & 9374 & 350332 & \cite{kotkov2022tag}\\ \hline
\end{tabular}
\caption{Datasets metadata and references. \label{tbl:datasets-metadata}}
\end{table}

\textbf{Experiment Setup:} For each experiment on real-life datasets, we first partition the data randomly dividing the set of users into 80\% of users for training and 20\% of users for testing. Within the set of training users, we further partition into 90\% for inference and 10\% for validation. For the prior distribution over the user parameters, we experimented with 10 prior distributions, all normal distributions but with different means and standard deviations. For each method, we run inference on the inference set to obtain an item estimate $\beta$. For ranking metrics evaluation, we compute top-$K$ accuracy with respect to the reference ranking predetermined by average ratings (after removing items with very high average ratings but receive very few ratings). For AUC and log-likelihood metrics, we choose the prior distribution over $\theta$ by evaluating the log-likelihood on the validation set. The prior distribution corresponding to the high validation log-likelihood is used to evaluate log-likelihood on the test set.

\newpage
\textbf{Python implementation.} For readers reading this paper online, we also include here the python implementation of our spectral algorithm.

\begin{lstlisting}[language=Python,basicstyle=\tiny]
import numpy as np
from scipy.sparse import csc_matrix

INVALID_RESPONSE = -99999

def construct_markov_chain_accelerated(X, lambd=0.1):
    m, _ = X.shape
    D = np.ma.masked_equal(X, INVALID_RESPONSE, copy=False)
    
    D_compl = 1. - D
    M = np.ma.dot(D, D_compl.T) # This computes Mij = sum_l Alj Ali Xli (1-Xlj)
    np.fill_diagonal(M, 0)
    M = np.round(M)
    
    # Add regularization
    M = np.where(np.logical_or((M != 0), (M.T != 0)), M+lambd, M)
    
    d = []
    # Construct a row stochastic matrix
    for i in range(m):
        di = max(np.sum(M[i, :]), 1)
        d.append(di)
        M[i, :] /= max(d[i], 1)
        M[i, i] = 1. - np.sum(M[i, :])

    d = np.array(d)
    return M, d
    

def spectral_estimate(X, max_iters=10000, lambd=1, eps=1e-6):
    """Estimate the hidden parameters according to the Rasch model, either for the tests' difficulties
    or the students' abilities. We follow the convention in Girth https://eribean.github.io/girth/docs/quickstart/quickstart/
    the response matrix X has shape (m, n) where m is the number of items and n is the number of users.
    The algorithm returns the item estimates.
    
    X: np.array of size (m, n) where missing entries have value INVALID_RESPONSE
    max_iters: int, maximum number of iterations to compute the stationary distribution of the Markov chain
    lambd: float, regularization parameter
    eps: tolerance for convergence checking
    
    """
    M, d = construct_markov_chain_accelerated(X, lambd=lambd)
    M = csc_matrix(M)
    
    m = len(A)        
    pi = np.ones((m,)).T
    for _ in range(max_iters):
        pi_next = (pi @ M)
        pi_next /= np.sum(pi_next)
        if np.linalg.norm(pi_next - pi) < eps:
            pi = pi_next
            break
        pi = pi_next
        
    pi = pi.T/d
    beta = np.log(pi)
    beta = beta - np.mean(beta)
    return beta

\end{lstlisting}

\end{document}